\documentclass[letterpaper]{article} 
\usepackage{aaai25}  
\usepackage{times}  
\usepackage{helvet}  
\usepackage{courier}  
\usepackage[hyphens]{url}  
\usepackage{graphicx} 
\usepackage{natbib}  
\usepackage{caption} 
\usepackage{algorithm}
\usepackage{algorithmic}
\usepackage{multirow}
\usepackage{booktabs}
\usepackage{color}

\usepackage{newfloat}
\usepackage{listings}
\DeclareCaptionStyle{ruled}{labelfont=normalfont,labelsep=colon,strut=off} 
\floatstyle{ruled}
\newfloat{listing}{tb}{lst}{}
\floatname{listing}{Listing}
\title{Frequency-Masked Embedding Inference: A Non-Contrastive Approach for Time Series Representation Learning}
\author{
    En Fu\textsuperscript{\rm 1},
    Yanyan Hu\textsuperscript{\rm 1,}\textsuperscript{\rm 2,}\thanks{Corresponding Author},
}
\affiliations{
    \textsuperscript{\rm 1}School of Intelligence Science and Technology, University of Science and Technology Beijing \\
    \textsuperscript{\rm 2}Institute of Artificial Intelligence, University of Science and Technology Beijing \\

    fuen@xs.ustb.edu.cn, huyanyan@ustb.edu.cn
}

\usepackage{bibentry}
\definecolor{DC}{RGB}{147,147,147}

\begin{document}

\maketitle

\begin{abstract}
Contrastive learning underpins most current self-supervised time series representation methods. The strategy for constructing positive and negative sample pairs significantly affects the final representation quality. However, due to the continuous nature of time series semantics, the modeling approach of contrastive learning struggles to accommodate the characteristics of time series data. This results in issues such as difficulties in constructing hard negative samples and the potential introduction of inappropriate biases during positive sample construction. Although some recent works have developed several scientific strategies for constructing positive and negative sample pairs with improved effectiveness, they remain constrained by the contrastive learning framework. To fundamentally overcome the limitations of contrastive learning, this paper introduces Frequency-masked Embedding Inference (FEI), a novel non-contrastive method that completely eliminates the need for positive and negative samples. The proposed FEI constructs 2 inference branches based on a prompting strategy: 1) Using frequency masking as prompts to infer the embedding representation of the target series with missing frequency bands in the embedding space, and 2) Using the target series as prompts to infer its frequency masking embedding. In this way, FEI enables continuous semantic relationship modeling for time series. Experiments on 8 widely used time series datasets for classification and regression tasks, using linear evaluation and end-to-end fine-tuning, show that FEI significantly outperforms existing contrastive-based methods in terms of generalization. This study provides new insights into self-supervised representation learning for time series. 
\end{abstract}

%
\begin{links}
    \link{Code/Appendix}{https://github.com/USTBInnovationPark/Frequency-masked-Embedding-Inference}
\end{links}

\section{Introduction}

Time series data is crucial in various industries' production processes and economic activities, serving as a foundational data format\cite{base1, base3}. Effective representation methods are essential for building highly transferable and generalizable pattern recognition modules, significantly reducing the optimization difficulty of deep models on small sample datasets\cite{base2, tst}. This has become a consensus in the deep learning community, particularly in the fields of computer vision (CV)\cite{surveyVit, visonMst, jepa} and natural language processing (NLP)\cite{bert, nlpCert, nlpConsert}.

However, in the field of time series analysis, self-supervised representation learning has not yet become to a community standard due to the insufficient generalization performance. Contrastive learning strategies have shown certain advantages and have become the foundational framework for much research in recent years\cite{ts2vec,tfc,infots}. Contrastive learning optimizes models by constructing positive and negative samples for anchor sample. Positive samples are created using data augmentation techniques to provide different views under similar semantics to the anchor sample, while negative samples are selected or constructed to have opposing semantics to the anchor sample. The main issue with this framework is the difficulty in defining appropriate positive and negative samples for time series data.

Due to the inherent continuity of time series, key characteristics such as trends and frequencies change continuously and cannot be fully enumerated. \textbf{This makes the boundaries between semantics in time series less clear compared to other data formats.} For example, it is easy to define a picture of a cat and a picture of a dog as opposites, but it is challenging to define whether a series with a 7-day cycle and that with a 6.5-day cycle are similar or opposite. They have differences but are not fundamentally opposed. Contrastive learning uses a discrete modeling approach to define absolute opposite semantics for all samples, which contradicts the inherent continuity of time series. 

Some methods\cite{infots,timesurl} have developed more reasonable and flexible strategies for constructing positive and negative samples to mitigate the issues mentioned above. However, they still operate within the discrete modeling framework of contrastive learning, failing to address the root cause of the problem. Therefore, this study aims to define the semantic differences in time series in a completely new way.

In this paper, a novel non-contrastive time series representation learning framework, Frequency-masked Embedding Inference(FEI), is proposed. Inspired by the successful application of Joint-Embedding Predictive Architecture (JEPA)\cite{jepa, ijepa} in CV, FEI establishes continuous relationships between different semantics in time series through the concept of embedding inference, constructing a continuous embedding space sensitive to frequency variations. FEI infers specific frequency sample directly in the embedding space using frequency masking prompts. Unlike contrastive learning, which models distinctions between semantics, FEI focuses on modeling the relationships between different semantics, achieving continuous modeling through a prompting strategy.
 
Overall, the contributions of this study include:

\begin{itemize}
\item This paper proposes FEI, a novel self-supervised representation learning framework for time series that eliminates the need for positive and negative sample pairs. Using frequency masking prompts, FEI performs embedding inference of different frequency bands in the embedding space, enabling continuous semantic modeling.
\item We validate the quality of the representation through linear evaluation and end-to-end fine-tuning experiments. The proposed FEI achieves new state-of-the-art performance across 8 benchmark datasets for classification and regression tasks requiring high-level representations.
\item This study demonstrates the feasibility and effectiveness of non-contrastive learning for time series self-supervised representation learning, providing new insights for further research in this area.
\end{itemize}

\section{Related Works}
\subsubsection{Time series representation learning.} In recent years, contrastive learning has become a fundamental paradigm for time series high-level representation learning. For example, in addition to learning series reconstruction representations in the embedding space, SimMTM\cite{simmtm} also uses contrastive loss to constrain the representation distance between positive and negative sample pairs. TimeDRL\cite{timedrl} uses CLS tokens to obtain high-level representations and generates positive and negative sample pairs at different stages of subspace mapping through gradient truncation to guide the high-level representation optimization of CLS tokens. COMET\cite{comet} constructs multi-level contrast constraints to achieve a stable high-level representation learning process. Due to the discrete modeling nature of the contrastive learning framework, the construction of positive and negative sample pairs can significantly impact the optimization results. Inappropriate pairing frequently occur in some self-supervised representation learning approaches based on contrastive learning. For this issue, TS2Vec\cite{ts2vec} proposing an enhanced contextual construction strategy to eliminate improper priors in positive and negative sample pairs. TimesURL\cite{timesurl} and TF-C\cite{tfc} both construct positive and negative samples from a frequency domain perspective; the former mixes the spectra of multiple samples to generate new augmented samples, while the latter constructs augmented samples through spectral masking and enhancement. In addition, both Literature \cite{R_contrastiveIssues1} and \cite{R_contrastiveIssues2} have also investigated the issues related to contrastive learning in time series modeling. However, these methods still operate within the contrastive learning framework and cannot fundamentally address the issues inherent in the discrete modeling approach of contrastive learning.

\subsubsection{Joint-Embedding Predictive Architecture.} JEPA \cite{jepa} is a self-supervised learning concept that introduces additional covariates in the latent space to guide the model in learning the relationships between multiple semantics. I-JEPA\cite{ijepa}, a successful application of this concept, demonstrates that human priors are unnecessary in the self-supervised process. This provides a potential solution to fundamentally address the issue of inappropriate biases introduced by sample augmentation in self-supervised representation learning. JEPA has garnered widespread attention in CV\cite{jepaCV, jepaPointCloud} and time series forecasting\cite{R_JEPA4Forecasting}. However, JEPA has not yet been applied to time series representation learning.

\section{Frequency-masked Embedding Inference}
\begin{figure}[t]
    \centering
    \includegraphics[width=1\linewidth]{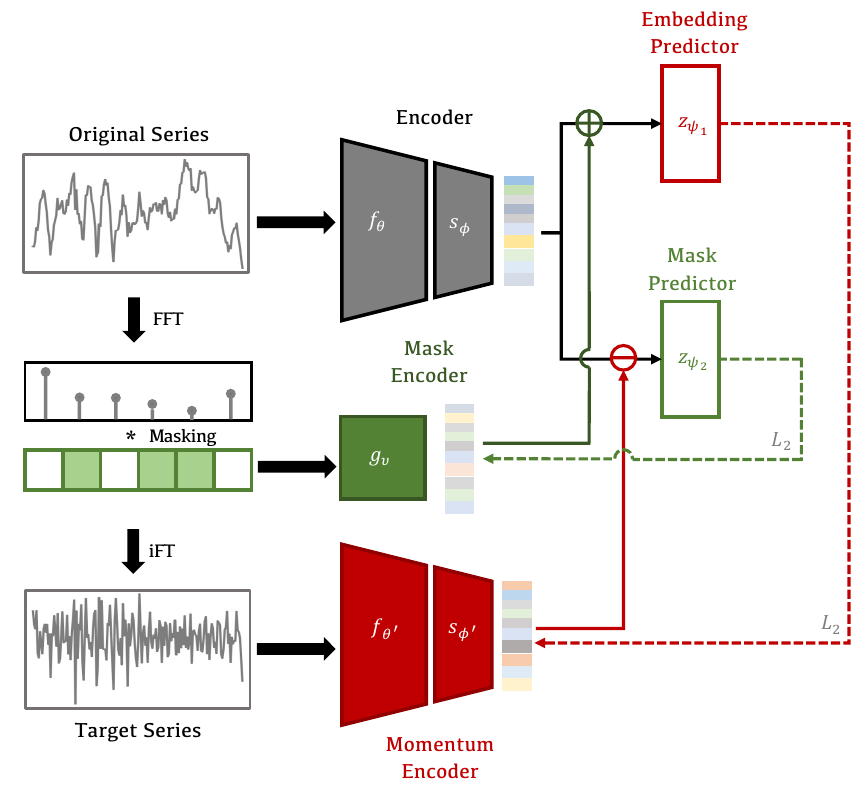}
    \caption{The overall structure of the proposed \textbf{FEI}. The original time series is fed into the encoder $f_\theta$ and a subspace projector $s_\phi$ to generate the original embedding. The target time series is constructed by applying random frequency masking, which is then fed into the momentum encoder—a smoothed copy of the original encoder updated via exponential moving average—to produce the target embedding. The goal of FEI is to enable the encoder to generate high-quality representation embeddings that can accurately infer the target embedding despite the presence of randomly masked frequency components(red dashed branch). Additionally, the representation embedding should also be capable of inferring mask embeddings by leveraging the differences between the target series and the original series (green dashed branch).}
    \label{fig:FEI-overall}
\end{figure}

The main architecture of the proposed FEI is illustrated in Figure \ref{fig:FEI-overall}. The core architecture of FEI consists of 2 sets of inference branches: \textbf{target embedding inference} based on mask prompts and \textbf{mask inference} based on target embedding prompts. The overall pre-training objective is to obtain a universal time series encoder $f_\theta$.

\subsection{Original Encoder}
Given the original time series $x\in \mathbf{R}^l$ with length $l$, the encoder $f_\theta$ generates an embedding vector $e\in \mathbf{R}^d$ for $x$, where $d$ represents the embedding dimension of $f_\theta$. This process does not impose any restrictions on the specific structure of encoder $f_\theta$. After obtaining the embedding vector, a projector $s_\phi$ serves as a relaxation factor to reduce the training complexity of the subsequent inference process, is used to obtain the subspace embedding $u\in \mathbf{R}^h$, where $h=d/2$. In our implementation, $s_\phi$ is implemented by a single linear layer.

\subsection{Momentum Encoder}
As FEI lacks explicit training constraints like those in contrastive learning architectures, directly performing embedding inference based on the single encoder may lead to representation collapse. A momentum encoder is an effective means to prevent this issue\cite{ijepa}. Therefore, FEI employs smoothly updated copies, $f_{\theta '}$ and $s_{\phi '}$, of the original encoder $f_{\theta}$ and projector $s_{\phi}$ as the target series encoder. These copies are updated using an exponential moving average and do not directly participate in gradient calculation. The update process is as follows: 
\begin{equation}
    \theta'_t = \alpha*\theta'_{t-1} + (1-\alpha)*\theta_{t}
    \label{eqa:momentum_update}
\end{equation}
where $\theta_t$ represents the weights of $f_{\theta}$ after the $t$-th gradient descent, $\theta'_t$ represents the weights of momentum encoder $f_{\theta '}$, and $\theta'_0$ = $\theta_0$. The same applies to subspace projector $s_{\phi '}$ and $s_{\phi}$. Similarly, the target series embedding $u'$ can be generated by $f_{\theta '}$ and $s_{\phi '}$.

\subsection{Frequency Masking} \label{sec:freq_masking}
To obtain the target series $x'$ corresponding to the original series $x$ while preserving its continuity, random frequency masking is employed. The original series $x$ undergoes a Fast Fourier Transform (FFT) to obtain its amplitude spectrum. A random mask $M\in {\{0, 1\}^n}$ with $k$ mask positions is then generated, where $n={\lfloor \frac{l}{2} \rfloor+1}$. The masked frequency component positions are marked as 1, and the unmasked positions are marked as 0.  This mask is applied to the amplitude spectrum of the original series, covering some of the frequency components, and the Inverse Fourier Transform (IFT) is then used to obtain the target time series $x'$ with masked frequencies.

It is important to note that the generation process of the frequency mask involves 2 steps. First, the number $k$ of masked frequency components is a random variable, with the masking ratio following a uniform distribution $U(\beta_1, \beta_2),0\leq\beta_1<\beta_2<1$. This means that each masking operation covers $\beta_1$ to $\beta_2$ of the total frequency components of the original series, providing sufficiently rich variation semantics during training. Then, $k$ frequency components are randomly selected as the final mask. This design avoids the issue of semantic uniformity caused by using the consistent masking ratio, while avoiding the introduction of inappropriate biases from any fixed masking pattern into the model training process.

\subsection{Mask Encoder}
To convert the frequency mask $M\in {\{0, 1\}^n}$ into a embedding vector that can be used as an embedding space prompt, we design a dedicated mask encoder to generate the mask embedding $m$ that adheres to the following principles: when no frequency components are masked, i.e., $M=\{0\}^n$, the mask embedding $m$ should also be 0. Additionally, the mean and variance of $m$ should remain relatively stable as $k$ varies. The mask encoder proposed is as follows: 
\begin{equation}
    m=g_\upsilon(M)=\frac{MW_{emb}}{\sqrt{k}}
\end{equation}
where $W_{emb} = \{w_1,...,w_n\}^T \in \mathbf{R}^{n\times h}$ constructs an embedding matrix, with each $w_i$ representing an embedding vector for a frequency component. $W_{emb}$ is randomly initialized from a normal distribution $N(0, 1)$ and updated during training.

\subsection{Embedding Inference}
The embedding inference part consists of 2 branches: the target embedding inference and the mask inference. Both branches use 1-layer MLP-based predictors, but differ in their prompt embeddings and merging methods. When inferring the target embedding, the mask embedding $m$ used to generate the target series is employed. Conversely, when inferring the mask embedding, the target embedding $u'$ is utilized. The computational process for this section is as follows: 
\begin{equation}
    \hat{u}' = z_{\psi_1}(u+D(m))
\end{equation}
\begin{equation}
    \hat{m} = z_{\psi_2}(D(u)-u')
\end{equation}
where $D(\cdot)$ represents the gradient detaching, $\hat{u}'$ is the inferred target embedding based on mask prompting, and $\hat{m}$ indicates the inferred mask embedding based on the target embedding. Gradient detaching is used to force the model to focus on different optimization aspects during the two inference processes. In the process of inferring the target embedding, the gradient calculation of the mask embedding $m$ is detached, and thus the mask encoder $g$ is not optimized. The encoder $f_\theta$ is driven to obtain better original embedding $u$ to complete the inference. Conversely, in the process of inferring the mask embedding, the gradient calculation of the original embedding $u$ is detached, and only the mask encoder is optimized to obtain a better mask embedding. This design clarifies the optimization objectives of the 2 branches and avoiding gradient conflicts during the training process.

The design of the embedding inference is the core module for the effectiveness of FEI and embodies the original intention behind FEI’s design. Firstly, we aim for the model to infer any changes in the series’s frequency band within the embedding space using appropriate prompts. Secondly, we aim for the model to infer the frequency differences between the original embedding and the frequency-masked target embedding. They jointly guide the model in understanding frequency variations in time series.

\subsection{Loss Function}
Based on the embedding inference results, the training loss is simply computed using the $L_2$ distance between the inferred target embedding $\hat{u}'$ and true target embedding $u'$, as well as between the inferred mask embedding $\hat{m}$ and the true mask embedding $m$, as follows:

\begin{equation}
    \mathcal{L}=\frac{1}{N} \sum\limits_{i=1}^{N} \|u'_i-\hat{u}'_i\|^2_2+ \|m_i-\hat{m}_i\|^2_2
\end{equation}

\section{Experiments}
This section reports the transfer performance of FEI on classification and regression downstream tasks, along with an ablation study analyzing the effectiveness of the FEI design. The full experiment results and more details can be found in Appendix.

\subsection{Dataset}
The dataset used in the experimental section is shown in Table \ref{tab:dataset_list}. For the pre-training phase, we use the commonly utilized SleepEEG dataset, which provides ample samples and is widely used for pre-training in various transfer learning methods \cite{tfc, simmtm}. For the downstream tasks,  8 publicly available datasets are utilized: 1) Gesture \cite{gesture}, 2) FD-B \cite{fdb}, 3) EMG \cite{EMG}, 4) EPI\cite{epi}, 5) HAR \cite{HAR}, 6) 128 UCR \cite{ucr}, 7) C-MAPSS \cite{cmapss}, and 8) Bearing \cite{bearing}. The first 6 datasets are commonly used for classification tasks, while the last 2 are typically used for regression tasks\cite{rul_base1, rul_base2}.
\begin{table}[t]
    \centering
    \begin{tabular}{c|cc}
         \toprule
         Task & Dataset & Freq.(Hz) \\
         \midrule
         Pre-training & SleepEEG &100\\
         \cmidrule{2-3}
         \multirow{6}*{Classification} & Gesture  &100\\
         & FD-B  &64k\\
         & EMG  &4k\\
         & EPI  &174\\
         & HAR  &50\\
         & 128 UCR  &-\\
         \cmidrule{2-3}
         \multirow{2}*{Regression} & C-MAPSS  &1\\
         & Bearing  &25.6k\\
         \bottomrule
    \end{tabular}
    \caption{The datasets used in experiments.}
    \label{tab:dataset_list}
\end{table}
\begin{center}
\end{center}
\subsection{Baselines}
To fully verify the representation advantages of FEI, we selected the state-of-the-art methods from recent years as baselines for comparison. These methods are: 1) TimesURL \cite{timesurl}, 2) SimMTM \cite{simmtm}, 3) InfoTS \cite{infots}, 4) TimeDRL \cite{timedrl}, 5) TF-C \cite{tfc}, and 6) TS2Vec \cite{ts2vec}. All these methods are representative of the current time series representation learning field and are based on contrastive learning frameworks. 

We reproduce these baselines using the official open-source code. To ensure a fair comparison of the generalization performance of pre-training methods, we use 1-D ResNet as the encoder network for all methods except TimesURL, with the embedding dimension set to 1024. The TimesURL method encounter an Out-of-Memory issue, preventing us from replacing it with a larger encoder network. However, we adjust the embedding dimension of TimesURL to be consistent with all other methods.

\subsection{Pre-training Setup}
The primary significance of representation learning is to enable the pre-trained model to be applicable to as many unknown downstream tasks as possible. Therefore, our experimental process fully adheres to this principle: \textbf{after completing pre-training, the encoder is directly transferred to different downstream datasets without any additional pre-training or hyperparameter adjustments for each dataset}. A single linear layer is used as the task-specific output layer. This approach fully verifies the differences in the representation quality of each method.

The basic setup of proposed FEI for the pre-training process follows SimMTM and TF-C. Additional configurations can be found in the Appendix.

\subsection{Task 1: Classification}
\subsubsection{Setup.}Time series classification is a key benchmark for evaluating representation learning algorithms. We perform linear evaluation on 6 commonly used datasets and conduct end-to-end fine-tuning on 5 datasets with limited training samples to validate the generalization capabilities of each method.

In the linear evaluation process, we freeze the encoder, and optimize only a linear classifier, with a maximum training iteration of 300 and an initial learning rate of 1e-4. In the end-to-end fine-tuning process, both the encoder and the classifier are optimized, with a maximum iteration of 100 and a smaller learning rate of 1e-5 to prevent overfitting. For both linear evaluation and end-to-end fine-tuning, all baseline encoders use the same hyperparameters. \textbf{The model with the lowest validation loss is used as the final test model} by early stopping for all datasets except for the 128 UCR dataset, which does not have a validation set division, so all models are tested directly after training ends.

We evaluate the accuracy, precision, recall, and F1 score of each method on classification tasks, with the accuracy results of linear evaluation shown in Table \ref{tab:linear_main_results_cls} and the fine-tuning results shown in Table \ref{tab:semi_main_results_cls}.

In these tables, "Rand. Init." represents the performance of a randomly initialized encoder used directly for downstream tasks without pre-training. 

\begin{table*}[t]
    \centering
    \begin{tabular}{c|c|ccccccc}
    \toprule
         Datasets   &  Rand. Init. &TS2Vec &TimeDRL    & TF-C  &   TimesURL    &   SimMTM  &   InfoTS  &   \textbf{FEI(Ours)} \\
    \midrule
         Gesture&   12.50 &   63.33    &50.00   &57.50    &   69.72    &74.17   &64.17    &\textbf{75.00} \\
         FD-B&   11.39 &   43.59    &40.63   &45.53    &   54.44    &60.74    &60.71    &\textbf{67.25} \\
         EMG&   46.34 &   \textbf{92.68}    &63.41   &78.05    &   \textbf{92.68}    &85.37    &87.80    &87.80 \\
         EPI&   19.79 &   96.41    &77.85   &85.75    &   95.42    &96.42   &96.27 &\textbf{96.84}  \\
         HAR&   36.51 &   78.91    &70.31    &67.56     &   79.10    &77.13     &78.35    &\textbf{79.54} \\
         128 UCR&   39.03 &   72.50   &61.61    &61.88     &   69.53    &75.34    &73.13     &\textbf{78.17}  \\
    \midrule
         Avg.&27.59 &72.66  &60.64  &64.38  &76.82  &78.19  &76.74  &   \textbf{80.77}\\
    \bottomrule
    \end{tabular}
    \caption{The \textbf{linear evaluation} accuracy(\%) of all methods on classification task.}
    \label{tab:linear_main_results_cls}
\end{table*}

\begin{table*}[t]
    \centering
    \begin{tabular}{c|c|ccccccc}
    \toprule
         Datasets   &  Rand. Init. &TS2Vec &TimeDRL    & TF-C  &   TimesURL    &   SimMTM  &   InfoTS  &   \textbf{FEI(Ours)} \\
    \midrule
         Gesture&   68.33 &   72.50    &73.33   &70.00    &   73.33    &76.67   &71.67    &\textbf{77.50} \\
         FD-B&   69.61 &   48.31    &47.97   &65.48    &   54.41    &63.49    &62.99    &\textbf{70.99} \\
         EMG&   95.12 &   78.05   &78.05   &92.68    &   73.17    &87.80    &\textbf{97.56}    &\textbf{97.56} \\
         EPI&   80.21 &   95.60    &94.05  &95.28    &   96.67    &96.22   &97.07 &\textbf{97.24}  \\
         128 UCR&   72.44 &   67.44   &63.36    &78.50     &   79.40    &80.42    &81.77     &\textbf{82.65}  \\
    \midrule
         Avg.&77.14 &73.62  &71.69  &80.86  &74.40  &81.05  &82.32  &   \textbf{85.82}\\
    \bottomrule
    \end{tabular}
    \caption{The \textbf{end-to-end fine-tuning} accuracy(\%) of all methods on small-sample classification datasets.}
    \label{tab:semi_main_results_cls}
\end{table*}

\subsubsection{Analysis.}From the results, it is evident that the proposed FEI achieves the best transfer performance in almost all cases. 
In linear evaluation scenarios, FEI achieves an average accuracy improvement of 2.15\% over existing methods on the compared datasets. In end-to-end fine-tuning, most models show varying degrees of improvement compared to linear evaluation results. The encoder trained by the proposed FEI still achieves the best average performance, with an average accuracy improvement of approximately 3.50\% on the 5 small-sample datasets compared to the best existing methods. The proposed FEI adopts an inference-based modeling approach for different frequency bands, fully leveraging the semantics of each frequency band in the pretraining dataset. This results in a broader learning corpus compared to contrastive-based methods, significantly enhancing the robustness of the encoder to variations in data frequency.


\subsection{Task 2: Regression}
\subsubsection{Setup.}Regression tasks are a classic type of time series task. In this paper, we use 2 equipment health status analysis datasets, C-MAPSS and Bearing, to analyze the transfer performance of FEI in regression tasks. These datasets consist of 4 and 3 sub-datasets, respectively, each using input signal to regress the Remaining Useful Life (RUL) ratio of the equipment. These datasets are commonly used for time series regression tasks in equipment health status analysis.

Moreover, the sampling frequency of these 2 datasets significantly differs from the pre-training dataset SleepEEG. The C-MAPSS dataset has a very low sampling frequency (1Hz), while the Bearing dataset has a very high sampling frequency (25.6Hz). This difference allows for a better evaluation of the model's generalization ability across different data characteristics.

The experimental settings for linear evaluation and end-to-end fine-tuning are the same as for classification tasks. Due to the smaller number of training samples in the Bearing dataset compared to the C-MAPSS dataset, end-to-end fine-tuning experiments are conducted on the Bearing dataset. Mean Squared Error (MSE) and Mean Absolute Error (MAE) are used as performance metrics, with results shown in Tables \ref{tab:linear_main_results_reg} and \ref{tab:semi_main_results_reg}, respectively. In the tables, all performance results are averaged across all sub-datasets in each dataset. 

\begin{table}[t]
    \centering
    \begin{tabular}{c|c|cc}
    \toprule
         Datasets& Methods& MSE&MAE\\
    \midrule
         \multirow{8}*{C-MAPSS}& Rand. Init.&0.0714&0.2439  \\
    \cmidrule{2-4}
                               & TS2Vec     &0.1172&0.2679  \\
                               & TimeDRL    &0.0644&0.2185  \\
                               & TF-C       &0.4499&0.5612  \\
                               & TimesURL   &0.1044&0.2566  \\
                               & SimMTM     &0.0599&0.2056  \\
                               & InfoTS     &0.0623&0.2145  \\
                               & \textbf{FEI(Ours)}   &\textbf{0.0584} &\textbf{0.1992}  \\
    \cmidrule{1-4}
         \multirow{8}*{Bearing}& Rand. Init.&0.5274&0.6705  \\
    \cmidrule{2-4}
                               & TS2Vec     &0.1268&0.1972  \\
                               & TimeDRL    &0.0919&0.1875  \\
                               & TF-C       &0.6022&0.6447  \\
                               & TimesURL   &0.0782&0.1813  \\
                               & SimMTM     &0.0959&0.1983  \\
                               & InfoTS     &0.0829&0.1905  \\
                               & \textbf{FEI(Ours)}   &\textbf{0.0744} &\textbf{0.1747} \\
    \bottomrule
    \end{tabular}
    \caption{The average \textbf{linear evaluation} results of all methods on regression task.}
    \label{tab:linear_main_results_reg}
\end{table}

\begin{table}[t]
    \centering
    \begin{tabular}{c|c|cc}
    \toprule
         Datasets& Methods& MSE&MAE\\
    \midrule
         \multirow{8}*{Bearing}& Rand. Init.&0.0728&0.1844  \\
    \cmidrule{2-4}
                               & TS2Vec     &0.1837&0.2649  \\
                               & TimeDRL    &0.1467&0.2885  \\
                               & TF-C       &0.6040&0.6446  \\
                               & TimesURL   &0.1641&0.2561  \\
                               & SimMTM     &0.0539&0.1397  \\
                               & InfoTS     &0.0732&0.1776  \\
                               & \textbf{FEI(Ours)}   &\textbf{0.0331} &\textbf{0.1150} \\
    \bottomrule
    \end{tabular}
    \caption{The \textbf{end-to-end fine-tuning} results of all methods on small-sample regression dataset.}
    \label{tab:semi_main_results_reg}
\end{table}

\subsubsection{Analysis.} As shown in the table, the proposed FEI method also achieves the best average performance in regression tasks. On the C-MAPSS and Bearing datasets, which exhibit significant semantic differences from the pre-training data, the TimesURL method improves the strategy for constructing positive and negative samples. Compared with TS2Vec, TimeDRL, and TF-C, its performance improves significantly. However, its MSE and MAE are still 9.31\% and 8.83\% higher than those of FEI on the C-MAPSS dataset, respectively. The FEI method completely abandons explicit negative sample constraints and instead establishes flexible correlations of frequency band features between samples. 


\subsection{Model Analysis}
\subsubsection{Ablation.}
To further explore the role of each module in the proposed FEI, an ablation analysis is conducted. We target the core modules of FEI for ablation, constructing 6 ablation models: 
\begin{itemize}
    \item $w/o$ \textbf{emb. infer.}, which removes the target embedding inference branch, retaining only the mask inference.
    \item  $w/o$ \textbf{mask prompt}, which removes the mask prompting for target series embedding inference.
    \item $w/o$ \textbf{momentum}, which removes the momentum encoder and directly uses the original encoder for encoding.
    \item $w/o$ \textbf{subspace}, which removes the subspace projector $s_\phi, s'_\phi$ and directly learns in the original embedding space.
    \item $w/o$ \textbf{mask infer.}, which removes the mask inference branch, retaining only the target embedding inference.
    \item $w/o$ \textbf{detach}, which removes the gradient detachment $D(\cdot)$ in Equations (3) and (4) training.
\end{itemize}

The linear evaluation precision of each ablation method on the EMG dataset is shown in Table \ref{tab:ablation_main}.
\begin{table}[t]
    \centering
    \begin{tabular}{c|c}
    \toprule
        Model   &   Prec.(\%)    \\
    \midrule
        \textbf{FEI}     &  \textbf{90.20}  \\
        $w/o$ emb. infer.     &  57.69 \textcolor{DC}{\small$\downarrow$32.51} \\
        $w/o$ mask prompt     &  62.75 \textcolor{DC}{\small$\downarrow$27.45} \\
        $w/o$ momentum     &  62.75 \textcolor{DC}{\small$\downarrow$27.50}\\
        $w/o$ subspace     &  51.60 \textcolor{DC}{\small$\downarrow$38.60}\\
        $w/o$ mask infer.     &  79.24 \textcolor{DC}{\small$\downarrow$10.96} \\
        $w/o$ detach     &  85.70 \textcolor{DC}{\small$\downarrow$4.50} \\
    \bottomrule
    \end{tabular}
    \caption{The ablation results of FEI on EMG dataset.}
    \label{tab:ablation_main}
\end{table}

Due to the design of the mask inference branch, remov-
ing the momentum encoder from FEI does not lead to sig-
nificant representation collapse. This is because trivial solu-
tions, where the encoder encodes all samples into the same
embedding vector, cannot minimize the loss of the mask in-
ference branch.

\begin{figure}[t]
    \centering
    \includegraphics[width=1\linewidth]{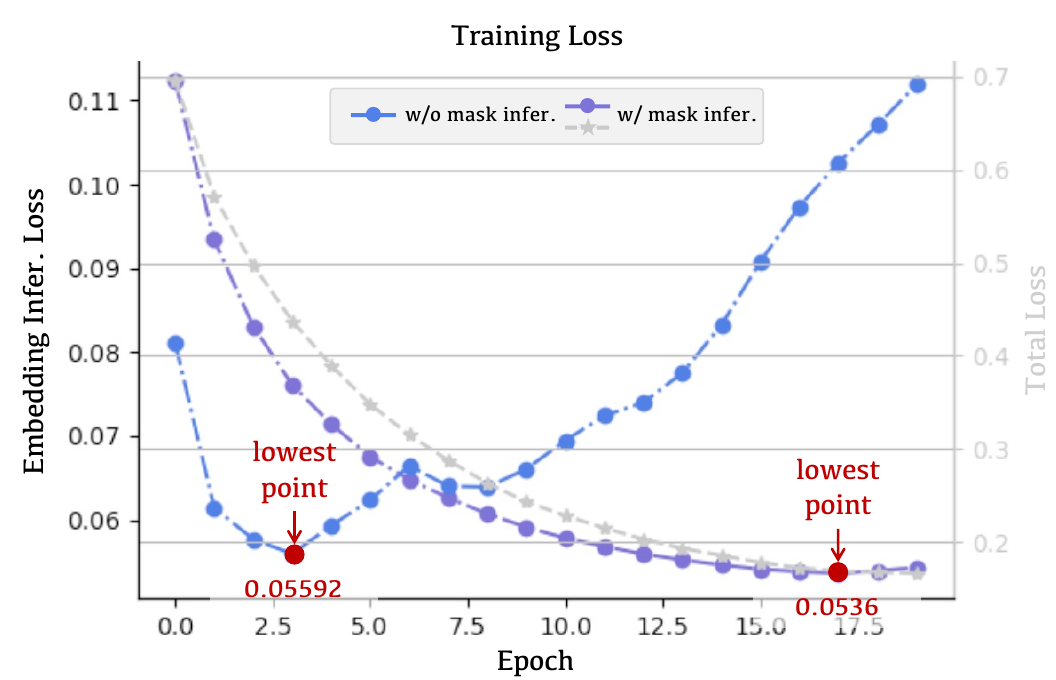}
    \caption{Comparison of training loss curves for the first 20 epochs $w/$ and $w/o$ mask inference.}
    \label{fig:training_loss_curve}
\end{figure}

Additionally, the mask inference stabilizes the FEI training process, as shown by the loss value comparison over the first 20 epochs in Figure \ref{fig:training_loss_curve}. The colored lines represent the loss of target embedding inference, while the gray lines represent the overall loss (target embedding inference + mask inference) with $w/$ mask infer. The mask inference provides a clearer optimization goal for the mask encoder $g_\upsilon$, and the smooth loss descent process indicates a flatter loss surface and better generalization\cite{flatness}.

\subsubsection{Visualization of embedding inference.}
\begin{figure}
    \centering
    \includegraphics[width=1\linewidth]{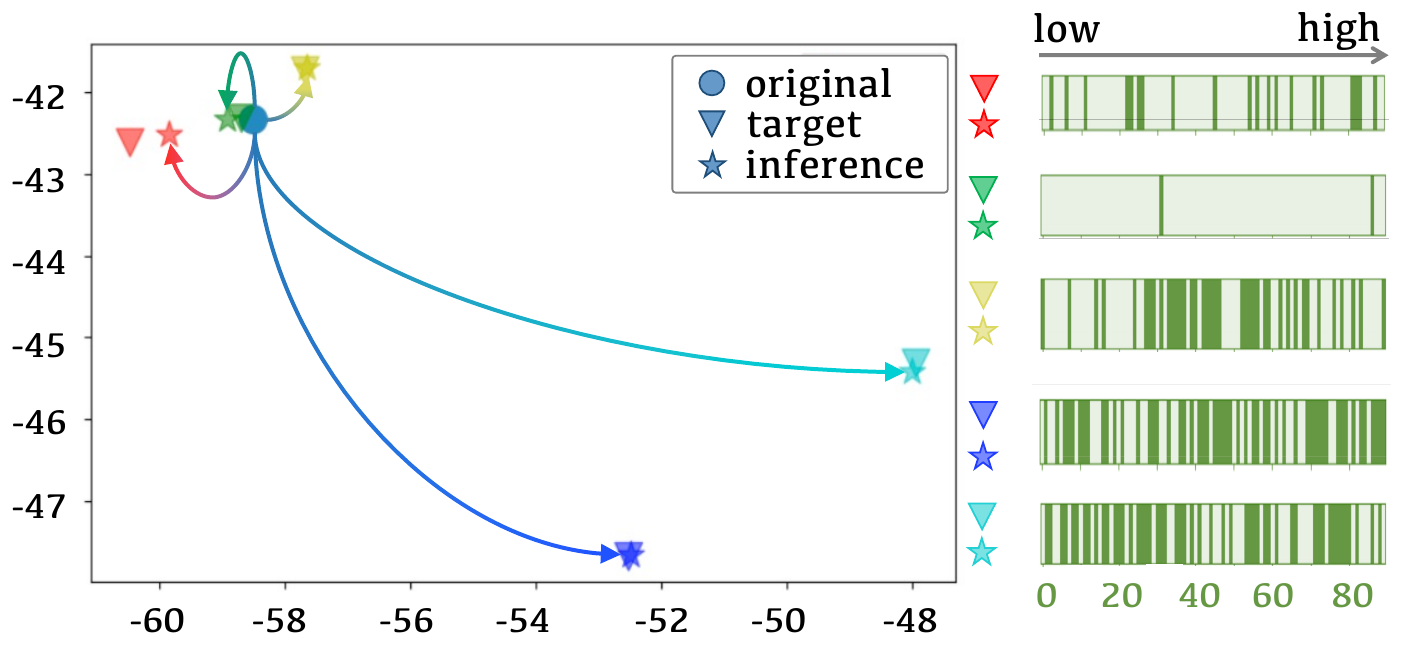}
    \caption{Visualization of embedding inference results on Gesture dataset. The left side shows the inference results, and the right side shows the masks used to construct the target series.}
    \label{fig:embedding_infer}
\end{figure}

Embedding inference is a core module of FEI. In this section, we select a original series sample from Gesture dataset and construct 5 representative target series using random masks. \textbf{It should be noted that FEI has never been trained on this dataset.} Using t-SNE\cite{tsne}, we visualize the original embedding and target embeddings after dimensionality reduction, as shown in Figure \ref{fig:embedding_infer}. The left side displays the relationship between the embeddings of the original series and the target series, with inverted triangles representing the target embeddings obtained directly using the original encoder $f_\theta$ and subspace projector $s_\phi$, and stars representing the inferred embeddings obtained by the embedding predictor $z_{\psi_1}$. The right side shows the 5 masks used to construct the target series, where the dark color represents the masked frequency component.
From the figure, it can be observed that the green target series has the fewest masks, and its embedding result and inference are close to the original embedding. The red and yellow target series have masks mainly in the mid and high-frequency bands, with only a small amount in the low-frequency band, resulting in their embeddings being further from the original embedding. The cyan and blue target series have the highest masking ratio and are significantly farther from the original embedding. FEI accurately complete the embedding inference for previously unseen series. The modeling approach of FEI encourages the encoder to establish a \textbf{continual frequency-sensitive embedding space}, significantly enhancing its general representation ability. Additional visualizations and details are available in the Appendix.

\subsubsection{Masking Strategies.}
To construct target sequences, we experiment with various masking strategies in the design of FEI, specifically:
\begin{itemize}
    \item Discrete Frequency Masking (DFM): Completely random frequency band masking strategy (used in FEI).
    \item Continuous Frequency Masking (CFM): Continuous frequency masking with random start and end points.
    \item Time-domain Masking (TDM): Random masking directly in the series's time domain.
\end{itemize}

We compare the performance differences of these masking strategies, and their accuracy results on the FD-B dataset are shown in Table \ref{tab:masking_strategy_main}. 

The key difference between frequency-domain masking(DFM, CFM) and time-domain masking(TDM) lies in the nature of the information loss they produce. Frequency-domain masking creates target series with soft information loss, where the semantic differences between the target and original series change continuously with the position and amount of frequency masking. In contrast, time-domain masking generates target series with hard information loss, causing semantic discontinuities in the time domain. We found that TDM leads to better transfer performance of FEI on datasets with the similar frequency (e.g., Gesture), but its performance significantly declines on datasets with different frequencies, particularly on the FD-B dataset with large frequency differences, which severely reduces FEI's generalization ability. 

Similarly, the SimMTM method, also based on time-domain masking, exhibits similar characteristics. As shown in Tables \ref{tab:linear_main_results_cls} and \ref{tab:linear_main_results_reg}, SimMTM demonstrates excellent transfer performance on datasets with the same frequency as the pre-training dataset (e.g., Gesture). However, its transfer performance significantly deteriorates on datasets with frequencies that differ greatly from the pre-training dataset (e.g., EMG, FD-B, Bearing). \textbf{This seems to suggest that time-domain masking, as a method for non-continuous semantic modeling, is more easily transferable to data with the same frequency but struggles to train models with strong generalization ability.} This phenomenon provides new insights for selecting sample augmentation strategies (time-domain processing, frequency-domain processing) in time series modeling. This further emphasizes the importance of continuous semantic modeling methods for achieving robust time series representations. Additional results can be found in the Appendix.

\begin{table}[t]
    \centering
    \begin{tabular}{c|cc}
    \toprule
        Masking Strategy & FD-B & Gesture \\
    \cmidrule{1-3}
        DFM & \textbf{67.25} & 75.00\\
        CFM & 60.65 \textcolor{DC}{\small$\downarrow 6.60$} & 71.67 \textcolor{DC}{\small$\downarrow 3.33$}\\
        TDM & 54.02 \textcolor{DC}{\small$\downarrow13.23$} & \textbf{79.17} \textcolor{black}{\small$\uparrow 4.17$}\\
    \bottomrule
    \end{tabular}
    \caption{The main results of different masking strategies on FD-B and Gesture datasets.}
    \label{tab:masking_strategy_main}
\end{table}

\subsubsection{Sensitivity}
\begin{figure}
    \centering
    \includegraphics[width=1\linewidth]{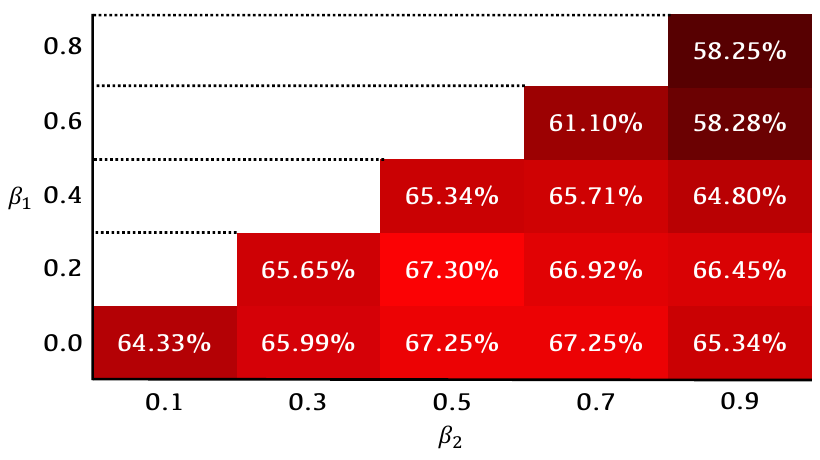}
    \caption{Accuracy results of FEI on the FD-B dataset under different masking ratios. The random masking ratio generated during each FEI training session lies between \(\beta_1\) and \(\beta_2\), as described in \textbf{Section Frequency Masking}.}
    \label{fig:sensitivity_masking_ratio}
\end{figure}

We further analyze the impact of different masking ratios on the performance of FEI. Figure \ref{fig:sensitivity_masking_ratio} shows the linear evaluation accuracy of FEI under different masking ratios on the FD-B dataset. Ultimately, we use $\beta_1=0.0$ and $\beta_2=0.7$.
More sensitivity analysis can be found in the Appendix.

\section{Conclusion}
This paper introduces FEI, a method for modeling the continuous semantic relationships of time series without the need for complex positive and negative sample pair construction. By inferring between different frequency-masked samples within the embedding space, FEI overcomes the limitations of contrastive learning. Experimental results demonstrate that the encoder trained by FEI achieves superior generalization performance compared to existing contrastive learning-based methods. 

\clearpage
\section{Acknowledgements}
This work is supported by the National Natural Science Foundation of China under Grants 62273038 and
U21A20483, and in part by the Innovative Talent Training Foundation for University of Science and Technology Beijing. Additionally, we sincerely thank the authors and open-source contributors who collected and processed the datasets used in this work (\citealt{sleepeeg}, \citealt{ucr}, \citealt{gesture}, \citealt{fdb}, \citealt{EMG}, \citealt{epi}, \citealt{HAR}, \citealt{cmapss}, \citealt{bearing}).
\bibliography{aaai25.bib}

\clearpage
\appendix \section{Dataset Details}
The details of the datasets used in the experiments are as follows:

\begin{itemize}
	\item \textbf{SleepEEG:} The SleepEEG dataset\cite{sleepeeg} consists of 153 whole-night sleep electroencephalogram (EEG) recordings, monitored using sleep cassette tapes. The data comes from 82 healthy subjects, with EEG signals sampled at 100 Hz. Each sample is associated with one of five sleep patterns/stages: Wakefulness (W), Non-Rapid Eye Movement (N1, N2, N3), and Rapid Eye Movement (REM). This dataset includes a mix of high- and low-frequency patterns and has been used in several studies as a pre-training dataset. We used the preprocessed version of the dataset, which was released by the TF-C\cite{tfc} researchers.
	\item \textbf{Gesture:} The Gesture\cite{gesture} dataset contains accelerometer measurements for 8 simple gestures, each varying based on the path of hand movement. The eight gestures include sliding the hand left, right, up, and down, waving in a clockwise or counterclockwise direction, waving in a square pattern, and waving in the shape of a right arrow. The classification labels correspond to these 8 different types of gestures. This dataset originates from the UCR dataset, and TF-C researchers have merged and cleaned it, making it widely used as a standalone small-sample dataset.
	\item \textbf{FD-B:} FD-B is a subset extracted from the FD\cite{fdb} dataset under condition B, which was collected from an electromechanical drive system that monitors the condition of rolling bearings and detects any damage. The data collected under different conditions is divided into 4 subsets, with parameters including rotational speed, load torque, and radial force. Each rolling bearing can be categorized into one of three classes: undamaged, internally damaged, and externally damaged. We used the preprocessed data provided by the TF-C researchers. 
	\item \textbf{EMG:} Electromyography(EMG) measures the electrical activity of muscles in response to nerve stimulation and can be used to diagnose certain muscular dystrophies and neuropathies. The EMG\cite{EMG} dataset consists of single-channel EMG recordings from the tibialis anterior muscle of three volunteers: one healthy, one with neuropathy, and one with myopathy. The data is sampled at a frequency of 4 kHz. Each patient, corresponding to their condition, represents a distinct classification category, with a total of three classes. We used the preprocessed data provided by the TF-C researchers.
	\item \textbf{EPI:} The Epilepsy(EPI)\cite{epi} dataset contains single-channel EEG measurements from 500 subjects. For each subject, brain activity was recorded for 23.6 seconds. The dataset was then divided and shuffled into 11,500 samples, each 1 second long, sampled at a frequency of 178 Hz. There are a total of 11,500 EEG samples classified into 2 categories, corresponding to epilepsy patients and normal patients. We also used the preprocessed data provided by the TF-C researchers.
	\item \textbf{HAR:} The HAR\cite{HAR} dataset includes recordings of 30 healthy volunteers performing six daily activities: walking, walking upstairs, walking downstairs, sitting, standing, and lying down. Each sample is labeled with one of these six activities. The dataset was obtained by measuring tri-axial acceleration and tri-axial angular velocity using wearable sensors on a smartphone, sampled at a frequency of 50 Hz. We also used the preprocessed dataset provided by the TF-C researchers.
	\item \textbf{128 UCR:} The UCR dataset\cite{ucr} includes a total of 128 time series classification datasets, making it one of the largest and most comprehensive benchmarks for time series classification. The dataset covers a wide range of fields, including healthcare, sports, and transportation, allowing for a thorough evaluation of algorithm classification performance. Although the dataset contains numerous subsets, the majority of these subsets have fewer than 500 training samples, making it suitable for evaluating transfer performance. In our experiments, we did not standardize the sample lengths; instead, we used the maximum sample length from each subset directly.
	\item \textbf{C-MAPSS:} The C-MAPSS\cite{cmapss} dataset, generated by NASA using the Commercial Modular Aero-Propulsion System Simulation, is used to predict the remaining useful life (from 100\% to 0\%) of an engine. It records key monitoring parameters of major components such as the combustion chamber and compressor during each simulated flight. The dataset contains 4 subsets FD001 to FD004, covering different operating conditions and causes of degradation. FD001 and FD003 are single-condition datasets, while FD002 and FD004 are multi-condition composite datasets. In our experiments, we selected the Low Pressure Compressor Outlet Temperature as the input variable to predict the remaining life of the engine.
	\item \textbf{Bearing:} The Bearing dataset\cite{bearing} originates from a real bearing machine setup, recording the vibration signals throughout the entire lifecycle of the bearings, from normal operation to eventual failure. It is also used to predict the remaining useful life. The dataset includes three different operational states with working speeds of 2100 rpm, 2250 rpm, and 2400 rpm. Vibration signals were collected using a DT9837 portable dynamic signal analyzer, with a sampling frequency of 25.6 kHz, a sampling interval of 1 minute, and each sampling duration of 1.28 seconds. We use a window size of 8192 to construct non-overlapping training and testing samples.
\end{itemize}

Detailed information about the datasets used in this study is shown in Table \ref{tab:dataset_details}. The sample lengths for the Gesture, FD-B, EMG, EPI, and HAR datasets listed in the table correspond to the lengths of the original dataset. In our experiments, the sample length used for all these datasets is 178, following the protocols established by TF-C\cite{tfc} and SimMTM\cite{simmtm}.
\begin{table*}[t]
	\centering
	\begin{tabular}{c|cc|ccc|ccc}
		\toprule
		\textbf{Task} & \textbf{Dataset} & \textbf{Sub-datasets} & \textbf{Train} &\textbf{Val.} &\textbf{Test} & \textbf{Length} & \textbf{Freq.(Hz)} & \textbf{Classes}  \\
		\midrule
		Pre-training& SleepEEG &-& 371,055 & 107,730 & - & 200   &100 & 5 \\
		\cmidrule{1-9}
		\multirow{6}{*}{Classification}& Gesture &-& 320 & 120 & 120 & 178   & 100 & 8 \\
		& FD-B &-& 60 & 21 & 13,559 & 5,120   &64k & 3 \\
		& EMG &-& 122 & 41 & 41  & 1,500   &4k & 3 \\
		& EPI &-& 60 & 20 & 11,420 & 178   &178 & 2 \\
		& HAR &-& 10,299 & 1,471 & 2,947 & 206   &50 & 6 \\
		& 128 UCR &-& 16$\sim$8,926 & - & 20$\sim$16,800 & 15$\sim$2,844   & - & 2$\sim$60 \\
		\cmidrule{1-9}
		\multirow{7}{*}{Regression}& \multirow{4}{*}{C-MAPSS} &FD001& 13,713 & 3,518 & 9,699 & 35   &1 & - \\
		& &FD002& 35,831 & 9,088 & 25,244  & 35   &1 & - \\
		& &FD003& 17,406 & 3,914 & 13,196 & 35   &1 & - \\
		& &FD004& 41,831 & 10,952 & 32,896 & 35   &1 & - \\
		\cmidrule{2-9}
		& \multirow{3}{*}{Bearing} &OC-A& 123 & 644 & 1,328 & 8,192   &25.6k & - \\
		& &OC-B& 491 & 644 & 3,656 & 8,192   &25.6k & - \\
		& &OC-C& 2,535 & 9,984 & 8,000 & 8,192   &25.6k & - \\
		\bottomrule
	\end{tabular}
	\caption{Detailed information about the datasets used in this study}
	\label{tab:dataset_details}
\end{table*}

\appendix \section{Pre-training Details of FEI}
The proposed FEI has few tunable hyperparameters, with the key performance-related settings listed in Table \ref{tab:hpparams}. During pre-training, an exponential decay strategy with a decay coefficient of 0.9 is used. The validation set from the pre-training dataset SleepEEG is used to monitor test loss and prevent overfitting, employing an early stopping mechanism with a patience value of 5 steps. The optimizer used is AdamW, with $\beta$ coefficients set to [0.9, 0.999]. 

\begin{table}[t]
	\centering
	\begin{tabular}{@{}ccp{4cm}@{}}
		\toprule
		\textbf{Hyperparameter} & \textbf{Value} & \textbf{Description} \\
		\toprule
		$epoch$ & 100 & The maximum pre-training epoch.  \\
		$bs$ & 512 & The batch size during pre-training. \\
		$lr$ & 0.0002 & Initial learning rate.\\
		$\alpha$   & $0.995$ & The momentum factor.\\
		$\beta_1$   & $0.0$ & Minimum length ratio of frequency masking.\\
		$\beta_2$   & $0.7$ & Maximum length ratio of frequency masking.\\
		\toprule
		
	\end{tabular}
	\caption{The hyperparameter configuration of FEI.}
	\label{tab:hpparams}
\end{table}

\appendix \section{Background and Baselines}
\subsection{Background} Currently, research on representation learning methods for time series is still in its early stages. Unlike other fields such as computer vision (CV) and natural language processing (NLP), it has yet to establish community standards. The benchmark experimental protocols for time series representation learning research are not fully established. By reviewing recent studies and their corresponding publicly available code, we found that existing experimental protocols mainly fall into two categories:
\begin{enumerate}
	\item Unsupervised/self-supervised pre-training is conducted on the target dataset, followed by either end-to-end fine-tuning or the construction of non-end-to-end machine learning algorithms on the same dataset. This constitutes an integrated approach, as seen in methods like TS2Vec \cite{ts2vec}, InfoTS \cite{infots}, TimesURL \cite{timesurl}, and TimeDRL \cite{timedrl}.
	\item Unsupervised/self-supervised pre-training is conducted on a pre-training dataset, followed by end-to-end fine-tuning on a target dataset that is different from the pre-training dataset. This is a two-stage approach, as seen in methods like TF-C\cite{tfc} and SimMTM\cite{simmtm}.
\end{enumerate}

Both of these approaches have certain limitations in validating the generalization performance of representation learning algorithms. First, Approach 1) cannot explicitly validate the generalization quality of the representations learned during the unsupervised representation learning stage because both pre-training and validation are confined to the same dataset. Approach 2) can only demonstrate that the representation learning algorithm helped the encoder find a good initial state during pre-training. However, it cannot validate whether the representations learned by the encoder meet the general requirements of the time series domain, as this approach lacks methods like linear evaluation that directly assess representation quality. Representation quality determines the potential of a method to train a general model with a larger encoder.

The evaluation of both representation quality and end-to-end fine-tuning performance is equally important in the field of time series representation learning, as it helps push the field towards generalization and standardization. The lack of consistency in experimental strategies has also led to significant fragmentation in the conclusions drawn by research in this field. One of the main objectives of this study is to develop a more comprehensive experimental strategy and benchmark to help unify research in this field. Therefore, in the implementation of our baseline, we consider the differences in implementing various approaches and conduct the experiments within a unified experimental framework as much as possible.

\subsection{Baselines}
Based on the above objectives, we conduct unified pre-training using 1D ResNet as the encoder for various baseline methods, including 1) TS2Vec, 2) TimeDRL, 3) TF-C, 4) TimesURL, 5) SimMTM, and 6) InfoTS, with the exception of TimesURL. Additionally, we design 2 types of experiments—linear evaluation and end-to-end fine-tuning on small samples—to comprehensively assess the performance differences among these methods. These methods are reproduced based on their publicly available code. The hyperparameter settings for all methods are kept consistent with those in their open-source code. Since TimeDRL has significantly different pre-training hyperparameters across different datasets, we select the hyperparameter settings that are similar across most datasets. The main hyperparameter settings for each baseline are shown in Table \ref{tab:hpparams_baseline1} and \ref{tab:hpparams_baseline2}.
\begin{table}[t]
	\centering
	\begin{tabular}{ccc}
		\toprule
		\textbf{Method} & \textbf{Hyperparameter} & \textbf{Value} \\
		\midrule
		\multirow{3}{*}{TS2Vec}
		&$epoch$ & 30\\
		&$bs$ & 32\\
		&$lr$ & 0.001\\
		\cmidrule{1-3}
		\multirow{5}{*}{TimeDRL}
		&$epoch$ & 30\\
		&$bs$ & 32\\
		&$lr$ & 0.0001\\
		&contrastive weight & 0.1\\
		&position embedding & learnable\\
		\cmidrule{1-3}
		\multirow{7}{*}{TF-C}
		&$epoch$ & 100\\
		&$bs$ & 256\\
		&$lr$ & 0.0003\\
		&temperature& 0.2\\
		&scale ratio& 1.5\\
		&jitter ratio& 2\\
		&max. seg.& 12\\
		\cmidrule{1-3}
		\multirow{7}{*}{TimesURL}
		&$epoch$ & 30\\
		&$bs$ & 128\\
		&$lr$ & 0.0001\\
		&$lmd$ & 0.01\\
		&temperature& 1.0\\
		&segment num.& 3\\
		&mask ratio& 0.05\\
		\bottomrule
		
	\end{tabular}
	\caption{The main hyperparameter settings of baselines (part 1).}
	\label{tab:hpparams_baseline1}
\end{table}

\begin{table}[t]
	\centering
	\begin{tabular}{ccc}
		\toprule
		\textbf{Method} & \textbf{Hyperparameter} & \textbf{Value} \\
		\midrule
		\multirow{7}{*}{SimMTM}
		&$epoch$ & 30\\
		&$bs$ & 128\\
		&$lr$ & 0.0001\\
		&length of masking & $1/3$\\
		&temperature& 0.2\\
		&positive num.& 3\\
		&mask ratio& 0.5\\
		\cmidrule{1-3}
		\multirow{9}{*}{InfoTS}
		&$epoch$ & 30\\
		&$bs$ & 512\\
		&$lr$ & 0.001\\
		&meta $lr$ & 0.01\\
		&$\beta$ & 0.5\\
		&meta $\beta$ & 0.1\\
		&$p_1$ & 0.2\\
		&$p_2$ & 0.0\\
		&mask mode& binomial\\
		\bottomrule
	\end{tabular}
	\caption{The main hyperparameter settings of baselines (part 2).}
	\label{tab:hpparams_baseline2}
\end{table}

\section{Full Results}
\subsection{Downstream Tasks} 
In this study, we validate performance on both classification and regression tasks. For classification tasks, the performance metrics include accuracy, precision, recall, and F1 score. For regression tasks, the performance metrics include MSE and MAE. The complete experimental results for both tasks are presented in Tables \ref{tab:full_results_cls_linear_eval}, \ref{tab:full_results_reg_linear_eval}, \ref{tab:full_results_cls_end2end}, and \ref{tab:full_results_reg_end2end}.All experiments were conducted on an Ubuntu 18.04 system with Python 3.8.15 and PyTorch 2.0.1, utilizing an Intel(R) Xeon(R) Silver 4210R CPU @ 2.40GHz, 64GB of RAM, and an Nvidia RTX 3090 GPU. We fixed the random seed at '2024' to ensure the reproducibility of all results.

\begin{table*}[t]
	\centering
	\begin{tabular}{c|c|cccc|c}
		\toprule
		Datasets & Methods & Accuracy(\%) & Precision(\%) & Recall(\%) & F1 Score(\%) & Mean(\%)\\
		\midrule
		\multirow{8}*{Gesture} & Rand. Init. & 12.50 & 12.50 & 1.56 &   2.78& 7.34 \\
		\cmidrule{2-7}
		& TS2Vec & 63.33 & 63.33 & 60.40 &   61.08  & 62.04\\
		& TimeDRL & 50.00 & 50.00 & 43.47 &   43.47  & 46.74\\
		& TF-C & 57.50 & 57.50 & 54.75 &   54.30  & 46.01\\
		& TimesURL & 69.72 & 69.72 & 64.66 &   65.60  & 67.43\\
		& SimMTM & 74.17 & 74.17 & 71.68 &   70.45  & 72.62\\
		& InfoTS & 64.17 & 64.17 & 58.78 &   60.32  & 61.86\\
		& \textbf{FEI} & \textbf{75.00} & \textbf{75.00} & \textbf{75.00} &   \textbf{72.54} & \textbf{74.39}\\
		\cmidrule{1-7}
		\multirow{8}*{FD-B} & Rand. Init. & 11.39 & 35.10 & 30.68 &   9.00& 21.54 \\
		\cmidrule{2-7}
		& TS2Vec & 43.59 & 31.91 & 28.79 &   28.83  & 33.28\\
		& TimeDRL & 40.63 & 42.09 & 37.96 &   37.03  & 39.43\\
		& TF-C & 45.53 & 33.36 & 48.51 &   20.91  & 37.08\\
		& TimesURL & 54.44 & 63.99 & 52.88 &   55.40  & 56.68\\
		& SimMTM & 60.74 & 69.98 & 57.96 &   60.94  & 62.41\\
		& InfoTS & 60.71 & 70.33 & 63.41 &   64.10  & 64.64\\
		& \textbf{FEI} & \textbf{67.25} & \textbf{75.09} & \textbf{66.58} &   \textbf{69.68}  & \textbf{69.68}\\
		\cmidrule{1-7}
		\multirow{8}*{EMG} & Rand. Init. & 46.34 & 33.33 & 15.45 &   21.11 & 16.52\\
		\cmidrule{2-7}
		& TS2Vec & \textbf{92.68} & 84.71 & \textbf{95.45} &   \textbf{88.22}  & \textbf{90.27}\\
		& TimeDRL & 63.41 & 48.30 & 43.77 &   45.65  & 50.28\\
		& TF-C & 78.05 & 68.44 & 74.49 &   70.18  & 72.79\\
		& TimesURL & \textbf{92.68} & 84.71 & \textbf{95.45} &   \textbf{88.22}  & \textbf{90.27}\\
		& SimMTM & 85.37 & 74.12 & 90.83 &   77.67  & 82.00\\
		& InfoTS & 87.80 & 66.67 & 59.02 &   62.54  & 78.69\\
		& \textbf{FEI} & 87.80 & \textbf{90.20} & 88.65 &   88.05  & 88.05\\
		\cmidrule{1-7}
		\multirow{8}*{EPI} & Rand. Init. & 19.79 & 50.00 & 9.89 &  16.52 & 24.05\\
		\cmidrule{2-7}
		& TS2Vec & 96.41 & 94.66 & 94.10 &   94.38  & 94.89\\
		& TimeDRL & 77.85 & 71.53 & 67.54 &   68.89  & 71.45\\
		& TF-C & 85.75 & 80.66 & 88.74 &   89.09  & 86.06\\
		& TimesURL & 95.42 & 93.62 & 92.23 &   92.90  & 93.54\\
		& SimMTM & 96.42 & \textbf{95.43} & 93.59 &   94.48  & 94.98\\
		& InfoTS & 96.27 & 94.82 & 93.62 &   94.21  & 94.73\\
		& \textbf{FEI} & \textbf{96.84} & 95.05 & \textbf{95.00} &   \textbf{95.02}  & \textbf{95.02}\\
		\cmidrule{1-7}
		\multirow{8}*{HAR} & Rand. Init. & 36.51 & 33.59 & 37.66 &   23.38 & 32.79\\
		\cmidrule{2-7}
		& TS2Vec & 78.91 & 78.40 & 81.44 &   77.76  & 79.13\\
		& TimeDRL & 70.31 & 69.03 & 73.21 &   64.88  & 69.36\\
		& TF-C & 67.56 & 66.14 & 80.49 &   64.82  & 69.75\\
		& TimesURL & 79.10 & 78.51 & 81.40 &   78.08  & 79.27\\
		& SimMTM & 77.13 & 76.33 & 80.05 &   75.57  & 77.27\\
		& InfoTS & 78.35 & 77.55 & \textbf{81.61} &  76.37  & 78.47\\
		& \textbf{FEI} & \textbf{79.54} & \textbf{79.06} & \textbf{81.61} &   \textbf{78.30}  & \textbf{79.63}\\
		\cmidrule{1-7}
		\multirow{8}*{128 UCR} & Rand. Init. & 39.03 & 35.07 & 26.93 &   26.86 & 31.97 \\
		\cmidrule{2-7}
		& TS2Vec & 72.50 & 69.31 & 71.15 &   67.61  & 70.15\\
		& TimeDRL & 61.61 & 59.76 & 60.21 &   57.49  & 59.77\\
		& TF-C & 61.88 & 58.68 & 63.56 &   57.09  & 60.31\\
		& TimesURL & 69.53 & 66.24 & 69.02 &   64.30  & 67.27\\
		& SimMTM & 75.34 & 73.37 & 74.96 &   72.75  & 74.11\\
		& InfoTS & 73.13 & 70.74 & 72.34 &   70.18  & 71.60\\
		& \textbf{FEI} & \textbf{78.17} & \textbf{76.37} & \textbf{77.94} &   \textbf{76.04}  & \textbf{77.13}\\
		\bottomrule
	\end{tabular}
	\caption{The full results of \textbf{linear evaluation} on 6 classification datasets.}
	\label{tab:full_results_cls_linear_eval}
\end{table*}

\begin{table*}[t]
	\centering
	\small
	\begin{tabular}{ccccccccccccccc}
		\toprule
		\multirow{3}*{Method} & \multicolumn{8}{c}{C-MAPSS}  & \multicolumn{6}{c}{Bearing} \\
		\cmidrule(r){2-9} \cmidrule(r){10-15}
		& \multicolumn{2}{c}{FD001}  & \multicolumn{2}{c}{FD002} &  \multicolumn{2}{c}{FD003}  & \multicolumn{2}{c}{FD004} &  \multicolumn{2}{c}{OC-A}  & \multicolumn{2}{c}{OC-B} & \multicolumn{2}{c}{OC-C} \\
		\cmidrule(r){2-9} \cmidrule(r){10-15}
		& MSE & MAE & MSE & MAE& MSE & MAE& MSE & MAE& MSE & MAE& MSE & MAE& MSE & MAE\\
		\cmidrule(r){1-1}\cmidrule(r){2-3}\cmidrule(r){4-5}\cmidrule(r){6-7}\cmidrule(r){8-9}\cmidrule(r){10-11}\cmidrule(r){12-13}\cmidrule(r){14-15}
		Rand. Init. & 0.065 & 0.233 & 0.094 & 0.282& 0.048 & 0.196& 0.079 & 0.265& 0.649 & 0.739& 0.575 & 0.689& 0.358 & 0.584 \\
		\cmidrule(r){1-1}\cmidrule(r){2-3}\cmidrule(r){4-5}\cmidrule(r){6-7}\cmidrule(r){8-9}\cmidrule(r){10-11}\cmidrule(r){12-13}\cmidrule(r){14-15}
		TS2Vec & 0.037 & 0.153 & \textbf{0.081} & \textbf{0.250}& 0.037 & 0.148& 0.233 & 0.405& 0.103 & \textbf{0.203}& 0.194 & 0.253& 0.084 & 0.136 \\
		TimeDRL& 0.046 & 0.179 & 0.096 & 0.284& 0.041 & 0.157& 0.075 & 0.254& 0.107 & 0.265& \textbf{0.074} & \textbf{0.177}& 0.054 & 0.102 \\
		TF-C& 0.353 & 0.495 & 0.268 & 0.499& 0.573 & 0.610& 0.606 & 0.642& 0.523 & 0.533& 0.472 & 0.543& 0.812 & 0.858 \\
		TimesURL& 0.046 & 0.174 & 0.096 & 0.274& 0.045 & 0.160& 0.173 & 0.336& \textbf{0.101} & {0.205}& 0.140 & 0.248& 0.034 & 0.110 \\
		SimMTM& 0.035 & 0.148 & 0.094 & 0.278& 0.035 & 0.142& 0.077 & 0.254& 0.105 & 0.265& 0.146 & 0.248& 0.037 & 0.083 \\
		InfoTS& 0.036 & 0.149 & 0.092 & 0.275& \textbf{0.033} & 0.160& 0.088 & 0.274& 0.105 & 0.263& 0.137 & 0.241& 0.026 & 0.068 \\
		\textbf{FEI}& \textbf{0.034} & \textbf{0.145} & 0.099 & 0.284& \textbf{0.033} & \textbf{0.132}& \textbf{0.068} & \textbf{0.236}& 0.104 & 0.261& {0.108} & {0.216}& \textbf{0.011} & \textbf{0.047} \\
		\bottomrule
	\end{tabular}
	\caption{The full results of \textbf{linear evaluation} on 2 regression datasets.}
	\label{tab:full_results_reg_linear_eval}
\end{table*}

\begin{table*}[t]
	\centering
	\begin{tabular}{c|c|cccc|c}
		\toprule
		Datasets & Methods & Accuracy(\%) & Precision(\%) & Recall(\%) & F1 Score(\%) & Mean(\%)\\
		\midrule
		\multirow{8}*{Gesture} & Rand. Init. & 68.33 & 68.33 & 63.87 &   64.34 & 66.22 \\
		\cmidrule{2-7}
		& TS2Vec & 72.50 & 72.50 & 71.59 &   71.72  & 72.08\\
		& TimeDRL & 66.67 & 66.67 & 61.33 &   61.63  & 64.08\\
		& TF-C & 70.00 & 70.00 & 65.32 &   66.73  & 68.01\\
		& TimesURL & 73.33 & 73.33 & 72.13 &   72.45  & 67.43\\
		& SimMTM & 76.67 & 76.67 & 75.28 &   74.03  & 75.66\\
		& InfoTS & 71.67 & 71.67 & 67.30 &   67.75  & 69.60\\
		& \textbf{FEI} & \textbf{77.50} & \textbf{77.50} & \textbf{75.35} &   \textbf{75.70} & \textbf{76.51}\\
		\cmidrule{1-7}
		\multirow{8}*{FD-B} & Rand. Init. & 69.61 & 77.53 & 70.23 &   72.27& 72.41 \\
		\cmidrule{2-7}
		& TS2Vec & 48.31 & 53.96 & 45.20 &   44.64  & 48.03\\
		& TimeDRL & 47.97 & 50.22 & 45.56 &   47.10  & 47.71\\
		& TF-C & 65.48 & 73.93 & 63.98 &   67.30  & 67.67\\
		& TimesURL & 54.41 & 63.90 & 53.33 &   56.45  & 57.02\\
		& SimMTM & 63.49 & 72.96 & 63.46 &   66.76  & 66.67\\
		& InfoTS & 62.99 & 72.57 & 65.12 &   66.87  & 66.89\\
		& \textbf{FEI} & \textbf{70.99} & \textbf{78.52} & \textbf{71.46} &   \textbf{74.29}  & \textbf{73.82}\\
		\cmidrule{1-7}
		\multirow{8}*{EMG} & Rand. Init. & 95.12 & 96.83 & 96.83 &   93.62 & 95.60\\
		\cmidrule{2-7}
		& TS2Vec & 78.05 & 69.27 & 69.74 &   68.17  & 71.31\\
		& TimeDRL & 78.05 & 64.15 & 85.35 &   65.70  & 73.31\\
		& TF-C & 92.68 & 94.53 & 90.63 &   92.31  & 92.54\\
		& TimesURL & 73.17 & 70.05 & 69.09 &   67.03  & 69.84\\
		& SimMTM & 87.80 & 71.37 & 92.24 &   72.64  & 81.01\\
		& InfoTS & \textbf{97.56} & \textbf{98.04} & \textbf{98.33} &   \textbf{98.14}  & \textbf{98.02}\\
		& \textbf{FEI} & \textbf{97.56} & \textbf{98.04} & 98.04 &   \textbf{98.14}  & 97.95\\
		\cmidrule{1-7}
		\multirow{8}*{EPI} & Rand. Init. & 80.21 & 50.00 & 40.11 &  44.51 & 53.71\\
		\cmidrule{2-7}
		& TS2Vec & 95.60 & 94.09 & 92.38 &   93.20  & 93.82\\
		& TimeDRL & 94.05 & 86.79 & 93.98 &   89.82  & 91.16\\
		& TF-C & 95.28 & 93.32 & 92.07 &   92.68  & 93.34\\
		& TimesURL & 96.67 & 96.01 & 93.86 &   94.89  & 95.36\\
		& SimMTM & 96.22 & 94.84 & 93.47 &   94.14  & 94.67\\
		& InfoTS & 97.07 & 95.32 & 95.43 &   \textbf{95.37}  & 95.80\\
		& \textbf{FEI} & \textbf{97.24} & \textbf{96.43} & 95.05 &   \textbf{95.72}  & \textbf{96.11}\\
		\cmidrule{1-7}
		\multirow{8}*{128 UCR} & Rand. Init. & 75.44 & 73.57 & 75.46 &   73.07 & 74.38 \\
		\cmidrule{2-7}
		& TS2Vec & 67.44 & 65.49 & 63.95 &   63.02  & 64.97\\
		& TimeDRL & 63.36 & 61.64 & 61.56 &   60.04  & 61.65\\
		& TF-C & 78.50 & 76.86 & 78.38 &   78.38  & 76.47\\
		& TimesURL & 79.41 & 77.79 & 78.84 &   77.21  & 78.31\\
		& SimMTM & 80.42 & 78.83 & 79.91 &   78.46  & 79.41\\
		& InfoTS & 81.78 & 80.28 & 81.31 &   79.93  & 80.82\\
		& \textbf{FEI} & \textbf{82.65} & \textbf{81.25} & \textbf{82.19} &   \textbf{80.95}  & \textbf{81.76}\\
		\bottomrule
	\end{tabular}
	\caption{The full results of \textbf{end-to-end fine-tuning} on 5 small-sample classification datasets.}
	\label{tab:full_results_cls_end2end}
\end{table*}

\begin{table*}[t]
	\centering
	\begin{tabular}{ccccccc}
		\toprule
		\multirow{3}*{Method} & \multicolumn{6}{c}{Bearing} \\
		\cmidrule(r){2-7}
		& \multicolumn{2}{c}{OC-A}  & \multicolumn{2}{c}{OC-B} & \multicolumn{2}{c}{OC-C} \\
		\cmidrule(r){2-7}
		& MSE & MAE & MSE & MAE& MSE & MAE\\
		\cmidrule(r){1-1}\cmidrule(r){2-3}\cmidrule(r){4-5}\cmidrule(r){6-7}
		Rand. Init. & 0.1618 & 0.3727 & 0.0449 & 0.1292& 0.0116 & 0.0513 \\
		\cmidrule(r){1-1}\cmidrule(r){2-3}\cmidrule(r){4-5}\cmidrule(r){6-7}
		TS2Vec &0.1560 &0.3378 &0.3561 &0.3629 &0.0394 &0.0939 \\
		TimeDRL &0.2312 &0.3907 &0.1626 &0.3055 &0.0463 &0.1694 \\
		TF-C &0.5286 &0.5327 &0.4717 &0.5429 &0.8119 &0.8583 \\
		TimesURL &0.1615 &0.3443 &0.2646 &0.3051 &0.0663 &0.1190 \\
		SimMTM &0.1174 &0.2352 &0.0265 &0.1317 &0.0178 &0.0522 \\
		InfoTS &0.1740 &0.3563 &0.0294 &0.1105 &0.0161 &0.0662 \\
		\textbf{FEI} &\textbf{0.0631} &\textbf{0.1848} &\textbf{0.0256} &\textbf{0.1099} &\textbf{0.0107} &\textbf{0.0503} \\
		\bottomrule
	\end{tabular}
	\caption{The full results of \textbf{end-to-end fine-tuning} on Bearing datasets.}
	\label{tab:full_results_reg_end2end}
\end{table*}

\subsection{Ablation}
We have reported the precision results of 6 ablation models on the EMG dataset in the main text. The complete ablation results are presented in Table \ref{tab:full_results_ablation}.


The ablation results show that each module design of FEI has varying degrees of impact on FEI's generalization ability. Overall, the most significant influences are from target embedding inference, mask prompting, and momentum encoder. The target embedding inference is the direct means by which FEI controls the embedding space. The mask prompting is the basis for meaningful embedding inference in FEI, as it indicates the inference targets, i.e., the frequency masking positions. Due to the mask inference branch, removing the momentum encoder will not fully degrade FEI, but it remains crucial for target embedding inference. The other modules serve as enhancement modules for FEI, having a relatively smaller impact on performance but still being indispensable. Among these, the subspace projector acts as a relaxation factor, allowing FEI to perform inferences in a more achievable subspace rather than the original embedding space. It has a significant impact on the transfer results for some datasets (e.g., the EMG dataset show in Table \ref{tab:full_results_ablation}). The mask inference and target embedding inference jointly reinforce the learning process of FEI.

\begin{table*}[t]
	\centering
	\begin{tabular}{c|c|cccc}
		\toprule
		Datasets & Structure & Accuracy(\%) & Precision(\%) & Recall(\%) & F1 Score(\%)\\
		\midrule
		\multirow{7}*{Gesture} 
		& \textbf{FEI} & 75.00 & 75.00 & 75.00 &   72.54\\
		& $w/o$ emb. infer. & 63.33 \textcolor{DC}{$\downarrow 11.67$} & 63.33 \textcolor{DC}{$\downarrow 11.67$} & 60.54 \textcolor{DC}{$\downarrow 14.46$} &   60.84 \textcolor{DC}{$\downarrow 11.70$}\\
		& $w/o$ mask prompt & 38.33 \textcolor{DC}{$\downarrow 36.67$} & 38.33 \textcolor{DC}{$\downarrow 36.67$} & 37.65 \textcolor{DC}{$\downarrow 37.35$} &   37.30 \textcolor{DC}{$\downarrow 35.24$}\\
		& $w/o$ momentum & 45.00 \textcolor{DC}{$\downarrow 30.00$} & 45.00 \textcolor{DC}{$\downarrow 30.00$} & 45.00 \textcolor{DC}{$\downarrow 30.00$} &   42.71 \textcolor{DC}{$\downarrow 29.80$}\\
		& $w/o$ subspace & 71.67 \textcolor{DC}{$\downarrow 3.33$} & 71.67 \textcolor{DC}{$\downarrow 3.33$} & 68.50 \textcolor{DC}{$\downarrow 6.50$} &   68.25 \textcolor{DC}{$\downarrow 4.29$}\\
		& $w/o$ mask infer. & 75.00 \textcolor{DC}{$-$} & 75.00 \textcolor{DC}{$-$} & 75.00 \textcolor{DC}{$-$} &   71.53 \textcolor{DC}{$\downarrow 1.01$}\\
		& $w/o$ detach & 71.67 \textcolor{DC}{$\downarrow 3.33$} & 71.67 \textcolor{DC}{$\downarrow 3.33$} & 67.14 \textcolor{DC}{$\downarrow 7.86$} &  68.48 \textcolor{DC}{$\downarrow 4.06$}\\
		\cmidrule{1-6}
		\multirow{7}*{FD-B} 
		& \textbf{FEI} & 67.25 & 75.09 & 66.58 &   69.68\\
		& $w/o$ emb. infer. & 34.49 \textcolor{DC}{$\downarrow 32.76$} & 36.93 \textcolor{DC}{$\downarrow 38.16$} & 34.17 \textcolor{DC}{$\downarrow 32.41$} &   31.17 \textcolor{DC}{$\downarrow 38.51$}\\
		& $w/o$ mask prompt & 37.46 \textcolor{DC}{$\downarrow 29.79$} & 38.22 \textcolor{DC}{$\downarrow 36.87$} & 37.94 \textcolor{DC}{$\downarrow 28.64$} &   30.84 \textcolor{DC}{$\downarrow 38.84$}\\
		& $w/o$ momentum & 54.28 \textcolor{DC}{$\downarrow 13.00$} & 57.21 \textcolor{DC}{$\downarrow 17.90$} & 57.21 \textcolor{DC}{$\downarrow 9.4$} &   51.14 \textcolor{DC}{$\downarrow 18.50$}\\
		& $w/o$ subspace & 59.00 \textcolor{DC}{$\downarrow 8.25$} & 66.69 \textcolor{DC}{$\downarrow 8.40$} & 56.60 \textcolor{DC}{$\downarrow 9.98$} &   59.53 \textcolor{DC}{$\downarrow 10.15$}\\
		& $w/o$ mask infer. & 65.15 \textcolor{DC}{$\downarrow 2.10$} & 73.15 \textcolor{DC}{$\downarrow 1.94$} & 65.17 \textcolor{DC}{$\downarrow 1.41$} &   68.23 \textcolor{DC}{$\downarrow 1.45$}\\
		& $w/o$ detach & 63.04 \textcolor{DC}{$\downarrow 4.21$} & 72.06 \textcolor{DC}{$\downarrow 3.03$} & 64.13 \textcolor{DC}{$\downarrow 2.45$} &  66.22 \textcolor{DC}{$\downarrow 3.46$}\\
		\cmidrule{1-6}
		\multirow{7}*{EMG} 
		& \textbf{FEI} & 87.80 & 90.20 & 88.65 &   88.05\\
		& $w/o$ emb. infer. & 34.49 \textcolor{DC}{$\downarrow 32.76$} & 36.93 \textcolor{DC}{$\downarrow 38.16$} & 34.17 \textcolor{DC}{$\downarrow 32.41$} &   31.17 \textcolor{DC}{$\downarrow 38.51$}\\
		& $w/o$ mask prompt & 82.93 \textcolor{DC}{$\downarrow 4.87$} & 62.75 \textcolor{DC}{$\downarrow 27.45$} & 55.31 \textcolor{DC}{$\downarrow 33.34$} &   58.73 \textcolor{DC}{$\downarrow 29.32$}\\
		& $w/o$ momentum & 82.93 \textcolor{DC}{$\downarrow 4.90$} & 62.75 \textcolor{DC}{$\downarrow 27.50$} & 55.80 \textcolor{DC}{$\downarrow 32.90$} &   58.87 \textcolor{DC}{$\downarrow 29.20$}\\
		& $w/o$ subspace & 68.29 \textcolor{DC}{$\downarrow 19.51$} & 51.60 \textcolor{DC}{$\downarrow 38.60$} & 45.41 \textcolor{DC}{$\downarrow 43.24$} &   48.25 \textcolor{DC}{$\downarrow 39.80$}\\
		& $w/o$ mask infer. & 85.37 \textcolor{DC}{$\downarrow 2.43$} & 79.24 \textcolor{DC}{$\downarrow 10.96$} & 89.44 \textcolor{black}{$\uparrow 0.79$} &   82.63 \textcolor{DC}{$\downarrow 5.42$}\\
		& $w/o$ detach & 87.80 \textcolor{DC}{$-$} & 85.70 \textcolor{DC}{$\downarrow 4.50$} & 91.07 \textcolor{black}{$\uparrow 2.42$} &  87.91 \textcolor{DC}{$\downarrow 0.14$}\\
		\cmidrule{1-6}
		\multirow{7}*{EPI} 
		& \textbf{FEI} & 96.84 & 95.05 & 95.00 &   95.02\\
		& $w/o$ emb. infer. & 94.44 \textcolor{DC}{$\downarrow 2.40$} & 87.86 \textcolor{DC}{$\downarrow 7.38$} & 94.41 \textcolor{DC}{$\downarrow 0.59$} &   90.55 \textcolor{DC}{$\downarrow 4.47$}\\
		& $w/o$ mask prompt & 95.64 \textcolor{DC}{$\downarrow 1.20$} & 92.80 \textcolor{DC}{$\downarrow 2.25$} & 93.38 \textcolor{DC}{$\downarrow 1.62$} &   93.09 \textcolor{DC}{$\downarrow 1.93$}\\
		& $w/o$ momentum & 95.20 \textcolor{DC}{$\downarrow 1.60$} & 92.69 \textcolor{DC}{$\downarrow 2.40$} & 92.69 \textcolor{DC}{$\downarrow 2.30$} &   92.69 \textcolor{DC}{$\downarrow 2.30$}\\
		& $w/o$ subspace & 95.15 \textcolor{DC}{$\downarrow 1.69$} & 92.74 \textcolor{DC}{$\downarrow 2.31$} & 92.10 \textcolor{DC}{$\downarrow 2.90$} &   92.42 \textcolor{DC}{$\downarrow 2.60$}\\
		& $w/o$ mask infer. & 96.04 \textcolor{DC}{$\downarrow 0.80$} & 93.47 \textcolor{DC}{$\downarrow 1.58$} & 94.00 \textcolor{DC}{$\downarrow 1.00$} &   93.73 \textcolor{DC}{$\downarrow 1.29$}\\
		& $w/o$ detach & 95.12 \textcolor{DC}{$\downarrow 1.72$} & 94.04 \textcolor{DC}{$\downarrow 1.01$} & 91.28 \textcolor{DC}{$\downarrow 3.72$} &  92.57 \textcolor{DC}{$\downarrow 2.45$}\\
		\cmidrule{1-6}
		\multirow{7}*{HAR} 
		& \textbf{FEI} & 79.54 & 79.06 & 81.61 &   78.30\\
		& $w/o$ emb. infer. & 75.06 \textcolor{DC}{$\downarrow  4.48$} & 74.36 \textcolor{DC}{$\downarrow  4.70$} & 76.86 \textcolor{DC}{$\downarrow  4.75$} &   73.90 \textcolor{DC}{$\downarrow  4.40$}\\
		& $w/o$ mask prompt & 68.44 \textcolor{DC}{$\downarrow 11.10$} & 67.27 \textcolor{DC}{$\downarrow 11.79$} & 70.53 \textcolor{DC}{$\downarrow 11.08$} &   66.54 \textcolor{DC}{$\downarrow 11.76$}\\
		& $w/o$ momentum & 74.92 \textcolor{DC}{$\downarrow  4.60$} & 74.08 \textcolor{DC}{$\downarrow  5.00$} & 76.83 \textcolor{DC}{$\downarrow  4.80$} &   73.57 \textcolor{DC}{$\downarrow  4.70$}\\
		& $w/o$ subspace & 78.25 \textcolor{DC}{$\downarrow 1.29$} & 77.45 \textcolor{DC}{$\downarrow 1.61$} & 81.26 \textcolor{DC}{$\downarrow 0.35$} &   76.44 \textcolor{DC}{$\downarrow 1.86$}\\
		& $w/o$ mask infer. & 78.01 \textcolor{DC}{$\downarrow  1.53$} & 77.28 \textcolor{DC}{$\downarrow  1.78$} & 79.83 \textcolor{DC}{$\downarrow 1.78$} &   76.60 \textcolor{DC}{$\downarrow 1.70$}\\
		& $w/o$ detach & 77.37 \textcolor{DC}{$\downarrow 2.17$} & 76.51 \textcolor{DC}{$\downarrow 2.55$} & 79.37 \textcolor{DC}{$\downarrow 2.24$} &  75.91 \textcolor{DC}{$\downarrow 2.39$}\\
		\bottomrule
	\end{tabular}
	\caption{The full results of ablation study.}
	\label{tab:full_results_ablation}
\end{table*}

\subsection{Masking Strategy}
We describe 3 masking strategies used in FEI to construct target series in the main text: Discrete Frequency Masking(DFM), Continuous Frequency Masking(CFM), and Time-domain Masking(TDM), with the full results listed in Table \ref{tab:full_results_mask}. 


\begin{table*}[t]
	\centering
	\begin{tabular}{c|c|cccc}
		\toprule
		Datasets & Structure & Accuracy(\%) & Precision(\%) & Recall(\%) & F1 Score(\%)\\
		\midrule
		\multirow{3}*{Gesture} 
		& DFM & 75.00 & 75.00 & 75.00 &   72.54\\
		& CFM & 71.67 \textcolor{DC}{$\downarrow 3.33$} & 71.67 \textcolor{DC}{$\downarrow 3.33$} & 67.68 \textcolor{DC}{$\downarrow 7.32$} &   68.32 \textcolor{DC}{$\downarrow 4.22$}\\
		& TDM & 79.17 \textcolor{black}{$\uparrow 4.17$} & 79.17 \textcolor{black}{$\uparrow 4.17$} & 77.18 \textcolor{black}{$\uparrow 2.18$} &   77.10 \textcolor{black}{$\uparrow 4.56$}\\
		
		\cmidrule{1-6}
		\multirow{3}*{FD-B} 
		& DFM & 67.25 & 75.09 & 66.58 &   69.68\\ 
		& CFM & 60.65 \textcolor{DC}{$\downarrow 6.60$} & 70.94 \textcolor{DC}{$\downarrow 4.15$} & 59.12 \textcolor{DC}{$\downarrow 7.46$} &   62.65 \textcolor{DC}{$\downarrow 7.12$}\\
		& TDM & 54.02 \textcolor{DC}{$\downarrow 13.23$} & 63.57 \textcolor{DC}{$\downarrow 11.52$} & 50.87 \textcolor{DC}{$\downarrow 15.71$} &   51.51 \textcolor{DC}{$\downarrow 18.17$}\\
		
		\cmidrule{1-6}
		\multirow{3}*{EMG} 
		& DFM & 87.80 & 90.20 & 88.65 &   88.05\\
		& CFM & 87.80 \textcolor{DC}{$-$} & 80.78 \textcolor{DC}{$\downarrow 36.67$} & 93.06 \textcolor{black}{$\uparrow 4.41$} &   84.56 \textcolor{DC}{$\downarrow 3.49$}\\
		& TDM & 73.17 \textcolor{DC}{$\downarrow 14.63$} & 69.85 \textcolor{DC}{$\downarrow 20.35$} & 81.06 \textcolor{DC}{$\downarrow 7.59$} &   73.63 \textcolor{DC}{$\downarrow 14.42$}\\
		
		\cmidrule{1-6}
		\multirow{3}*{EPI} 
		& DFM & 96.84 & 95.05 & 95.00 &   95.02\\
		& CFM & 96.39 \textcolor{DC}{$\downarrow 0.45$} & 95.23 \textcolor{black}{$\uparrow 0.18$} & 93.66 \textcolor{DC}{$\downarrow 1.34$} &   94.42 \textcolor{DC}{$\downarrow 0.60$}\\
		& TDM & 94.26 \textcolor{DC}{$\downarrow 2.58$} & 90.99 \textcolor{DC}{$\downarrow 4.06$} & 90.93 \textcolor{DC}{$\downarrow 4.07$} &   90.06 \textcolor{DC}{$\downarrow 4.06$}\\
		
		\cmidrule{1-6}
		\multirow{3}*{HAR} 
		& DFM & 79.54 & 79.06 & 81.61 &   78.30\\
		& CFM & 78.59 \textcolor{DC}{$\downarrow 0.95$} & 77.85 \textcolor{DC}{$\downarrow 1.21$} & 80.50 \textcolor{DC}{$\downarrow 1.11$} &   77.38 \textcolor{DC}{$\downarrow 0.92$}\\
		& TDM & 78.62 \textcolor{DC}{$\downarrow 0.92$} & 78.00 \textcolor{DC}{$\downarrow 1.06$} & 79.73 \textcolor{DC}{$\downarrow 1.88$} &   77.31 \textcolor{DC}{$\downarrow 0.99$}\\
		
		\bottomrule
	\end{tabular}
	\caption{The full results of different masking strategies.}
	\label{tab:full_results_mask}
\end{table*}

\subsection{Sensitivity}
\begin{figure}
	\centering
	\includegraphics[width=1\linewidth]{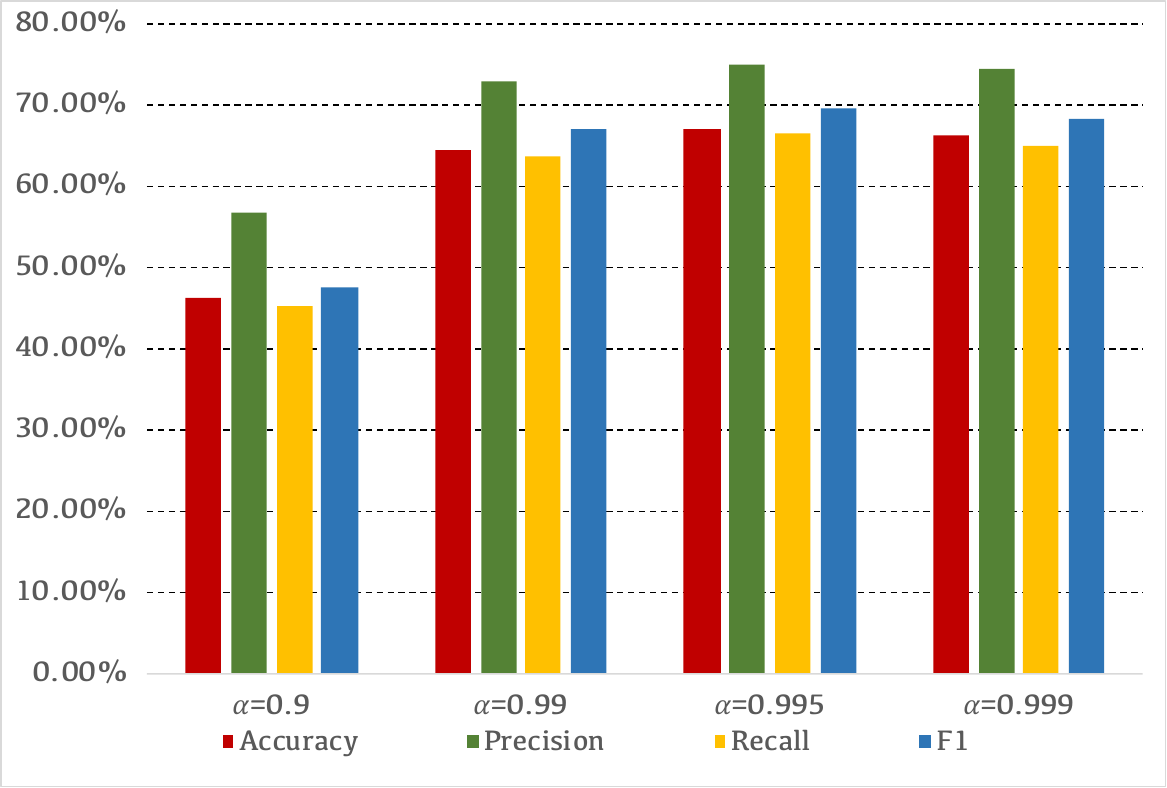}
	\caption{The linear evaluation results for varying momentum factors $\alpha$ on FD-B dataset.}
	\label{fig:momentum_results}
\end{figure}

We have reported FEI's sensitivity to masking ratio $\beta_1$and $\beta_2$. In addition, the momentum factor $\alpha$ determines the update speed of the momentum encoder, which is crucial for most methods based on momentum update strategies, such as BYOL\cite{byol} and I-JEPA\cite{ijepa}. Therefore, we further analyze the impact of the momentum update factor $\alpha$ on FEI. The results are shown in Figure \ref{fig:momentum_results}.The results indicate that the optimal range for $\alpha$ in FEI is between 0.99 and 0.999, within which sufficient representational generalization can be achieved. Ultimately, we use $\alpha=0.995$.

\subsection{Visualization}
\begin{figure*}
	\centering
	\includegraphics[width=1\linewidth]{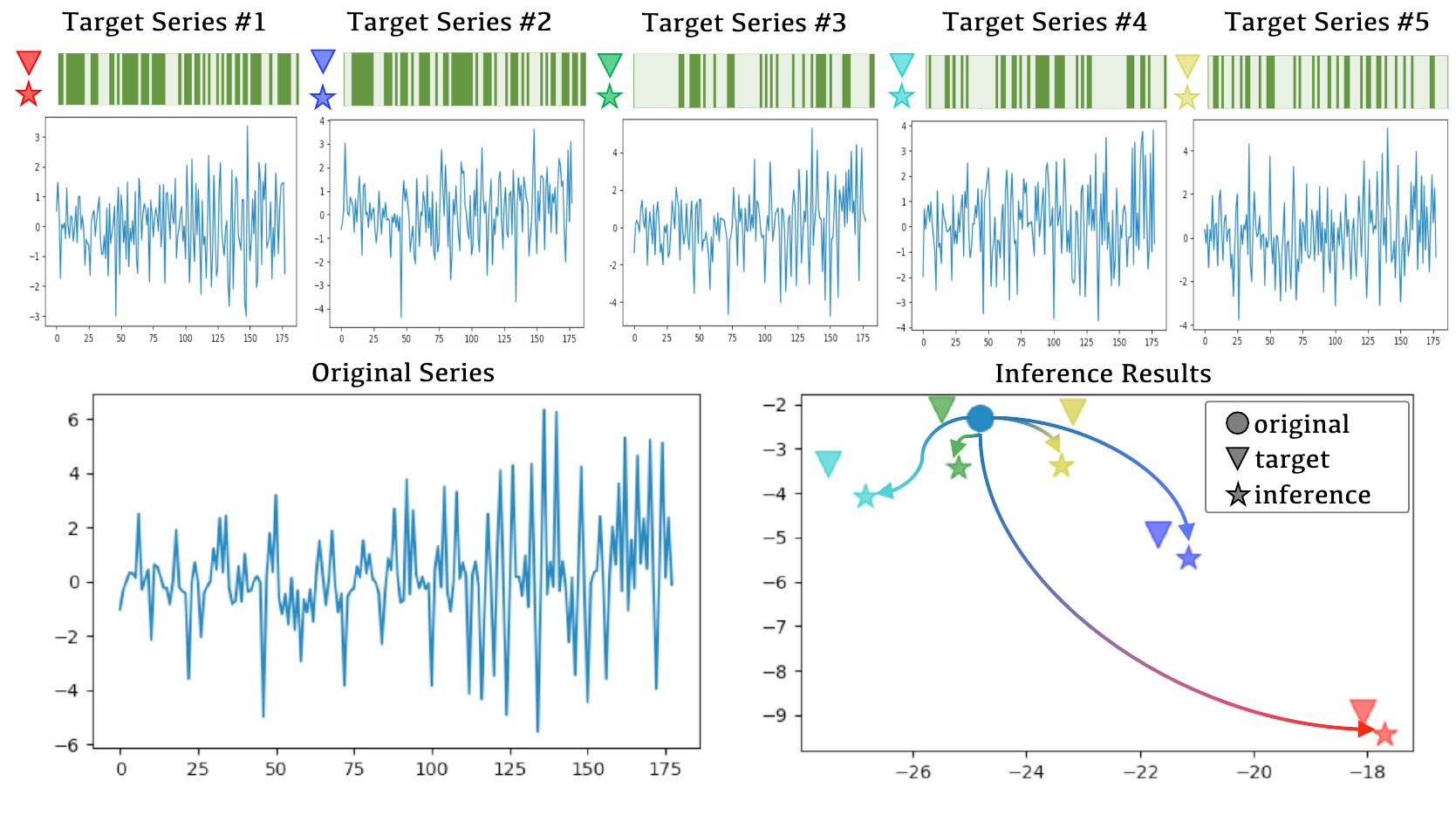}
	\caption{Visualization of target series and inference results on the \textbf{FD-B} dataset.}
	\label{fig:visual_fdb}
\end{figure*}

\begin{figure*}
	\centering
	\includegraphics[width=1\linewidth]{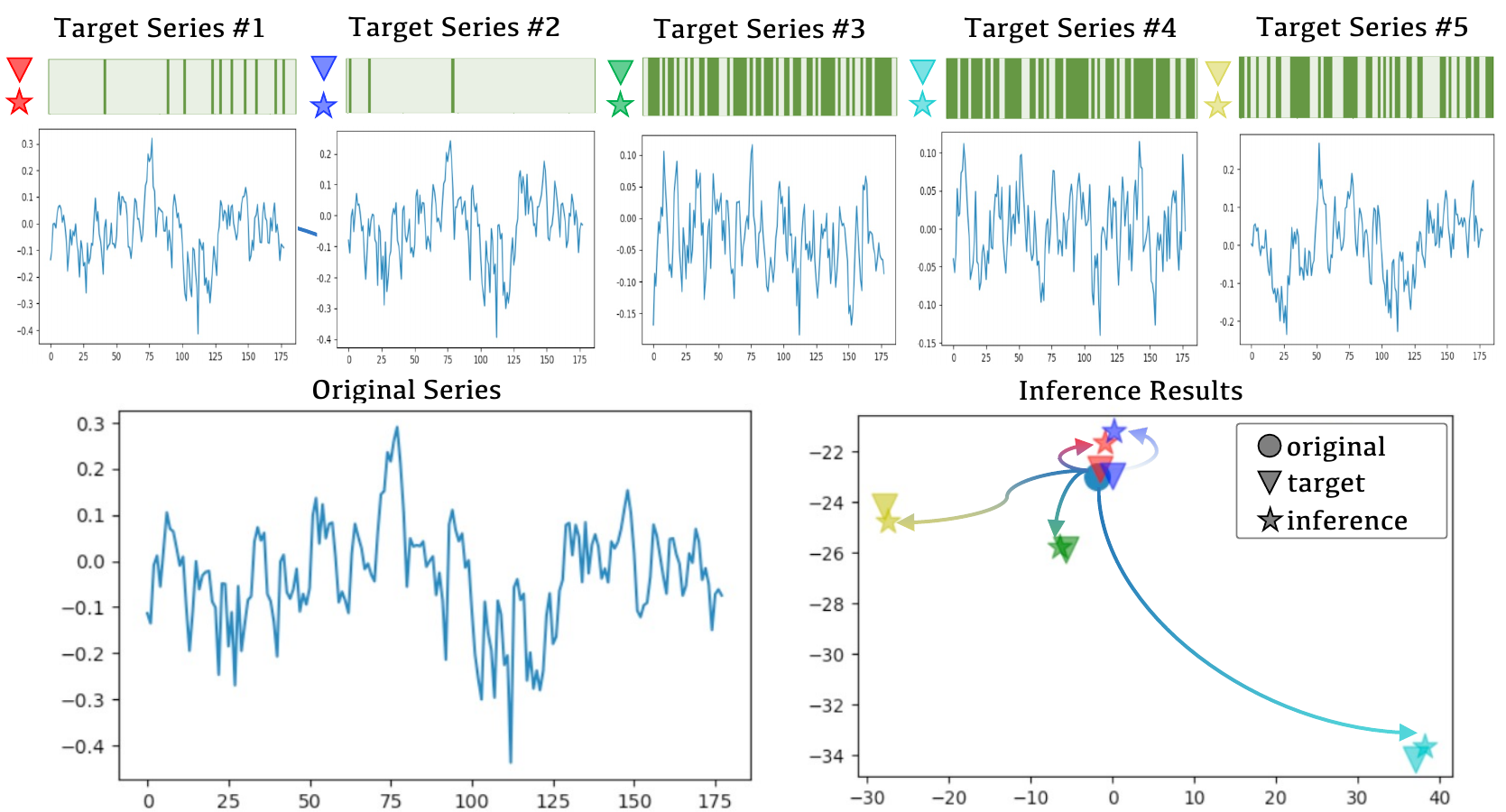}
	\caption{Visualization of target series and inference results on the \textbf{EMG} dataset.}
	\label{fig:visual_emg}
\end{figure*}

FEI demonstrates good sensitivity to changes in the frequencies of unseen samples. We have already shown the FEI's inference visualization on the unseen Gesture dataset in the main text. Here, we provide additional, more detailed visualizations on the \textbf{FD-B} and \textbf{EMG} datasets, which have significant frequency differences from the pre-training dataset SleepEEG, as shown in Figures \ref{fig:visual_fdb} and \ref{fig:visual_emg}.

In the figures, "Target Series \#n" represents the target series constructed from the original series using the frequency mask shown below. In the "Inference Results" section, the inverted triangle represents the true embedding of the target series, the star represents the inferred embedding of the target series obtained by FEI using the mask prompt and original series, and the circle represents the embedding of the original series. All embeddings are displayed using t-SNE dimensionality reduction.

Both sets of visualizations include low-frequency and high-frequency masking to varying degrees, and FEI accurately infers embeddings, even though these samples were never used to train FEI. Thanks to the modeling approach of embedding inference, FEI can generate meaningful embedding for time series sample based on its frequency characteristics, which is the source of FEI's strong generalization ability.

\section{Future Work}

The proposed FEI introduces a new modeling approach for self-supervised representation learning of time series. Through extensive experiments, we have validated and analyzed the effectiveness and superiority of FEI. However, advancing this field remains crucial, as developing a fully generalized temporal representation model holds significant importance for the entire time series analysis domain. Based on the current limitations of FEI and recent research progress, we offer some recommendations for future research.

FEI successfully achieves sample-level general representation modeling and has shown significant improvements in sequence-level downstream tasks such as classification and regression. However, we have not yet explored how this architecture could be applied to finer-grained modeling at the time-step level. This could be valuable for certain downstream tasks like point-to-point anomaly detection or stepwise time series prediction. Our findings suggest that continuous semantic modeling at the step level is beneficial for obtaining generalizable time series representation models. \textbf{This leads to our first suggestion for future work: exploring continuous semantic modeling frameworks at the time-step level, which may not be limited to the FEI architecture.}

The inherent diversity of time series data, which can describe a wide range of objects, results in significant differences in key features such as trends, cycles, and noise levels between different series. Thus, learning universal representations for time series remains a highly challenging field. Constructing a comprehensive, large-scale time series corpus that covers all possible objects of description is extremely difficult. This challenge also differentiates the design of self-supervised learning algorithms in this field from those in CV and NLP, which benefit from vast amounts of data. The frequency inference approach of FEI offers a new perspective for training general time series representation models with limited samples. By using frequency-domain processing methods such as frequency masking, as demonstrated in this paper, it is possible to construct a large number of new samples that are temporally continuous but semantically distinct from the original samples. These differences and relationships among the constructed samples can guide self-supervised learning. Our proposed FEI utilizes embedding inference to model semantic relationships, representing a practical application of this idea. Moving forward, we believe it is possible to develop representation learning models in the time series domain that achieve sufficient representational power with fewer pretraining samples. \textbf{Therefore, a second potential research direction is to explore frameworks for mining semantic relationships within time series, which can train general representation models based on a large number of constructible samples. This presents an alternative research path to building vast time series pretraining corpora.}

\clearpage
\end{document}